%% file: main.tex
\documentclass[sigconf]{acmart}

\newcommand{\final}{0}
\input{packages}

\input{macros}

\usepackage[ruled,linesnumbered]{algorithm2e}

\SetAlFnt{\small}
\SetAlCapFnt{\small}
\SetAlCapNameFnt{\small}
\SetAlCapHSkip{0pt}
\AtBeginDocument{%
  \providecommand\BibTeX{{%
    \normalfont B\kern-0.5em{\scshape i\kern-0.25em b}\kern-0.8em\TeX}}}

\setcopyright{acmlicensed}
\copyrightyear{2018}
\acmYear{2018}
\acmDOI{XXXXXXX.XXXXXXX}
\renewcommand\footnotetextcopyrightpermission[1]{}
\acmConference[]{}

\acmISBN{}
\renewcommand\footnotetextcopyrightpermission[1]{} 
\begin{document}

\title{Break-for-Make: Modular Low-Rank Adaptations for Composable Content-Style Customization}

\author{
Yu Xu\textsuperscript{1,2}, 
Fan Tang\textsuperscript{1,2},
Juan Cao\textsuperscript{1,2}, 
Yuxin Zhang\textsuperscript{3,2}, 
Oliver Deussen\textsuperscript{4}, 
Weiming Dong\textsuperscript{3,2},
Jintao Li\textsuperscript{1}, \\
Tong-Yee Lee\textsuperscript{5}
}
\affiliation{%
  \institution{
$^{1}$Institute of Computing Technology, Chinese Academy of Sciences\\
$^{2}$University of Chinese Academy of Sciences
$^{3}$Institute of Automation, Chinese Academy of Sciences\\
$^{4}$University of Konstanz
$^{5}$National Cheng-Kung University\\
}
\country{}
}

\email{{xuyu21b,tangfan,caojuan,jtli}@ict.ac.cn, {zhangyuxin2020,weiming.dong}@ia.ac.cn,}
\email{oliver.deussen@uni-konstanz.de, tonylee@mail.ncku.edu.tw}

\renewcommand{\shortauthors}{Xu, et al.}


\input{Sections/0_abstract}


\begin{teaserfigure}
  \centering
   \includegraphics[width=1.0\linewidth]{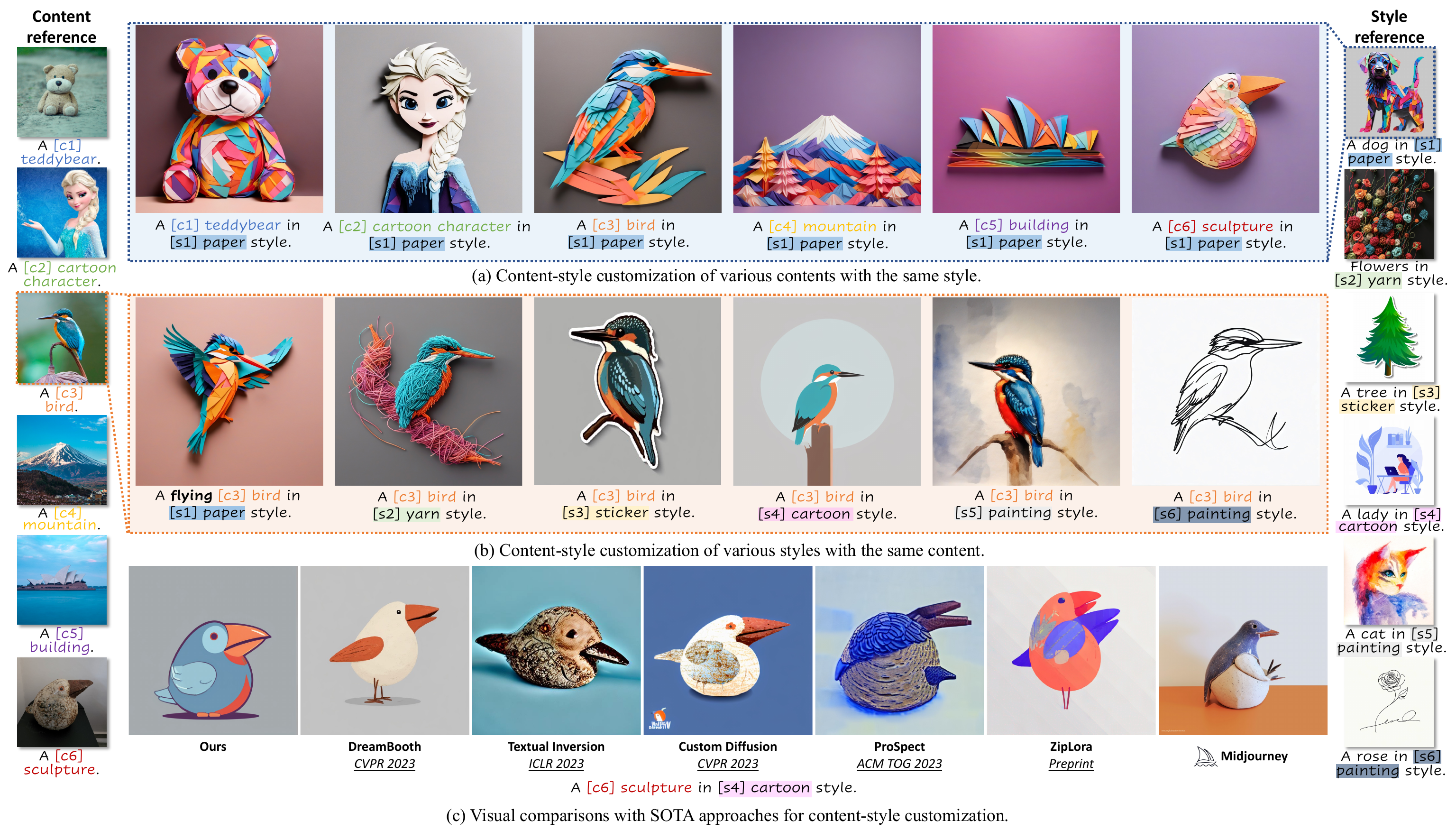}
   \caption{By separately learning content and style in ``partly learnable projection'' (\textbf{PLP}), our method is able to generate images of customized content and style aligned with various prompts while successfully disentangling content and style and maintaining high fidelity of them.}
  \label{fig:head}
\end{teaserfigure}

\maketitle

\input{Sections/1_intro}

\input{Sections/2_relate}
\input{Sections/3_method}

\input{Sections/4_experiments}

\newpage
\bibliographystyle{ACM-Reference-Format}
\balance
\bibliography{BfM}
 
\end{document}

%% file: packages.tex
\usepackage{color, soul}
\usepackage{booktabs}
\usepackage{makecell}
\usepackage{verbatim}
\usepackage{tabularray}
\usepackage{hyperref}
\usepackage{multirow}
\usepackage{xcolor}
\usepackage{ifthen}
\usepackage{microtype}
\usepackage{amsmath}
\usepackage{graphicx}
\usepackage{amsfonts}
\usepackage{array}
\usepackage[caption=false,font=normalsize,labelfont=sf,textfont=sf]{subfig}
\usepackage{stfloats}
\usepackage{url}
\usepackage{ragged2e}

\usepackage{caption}
\usepackage{subcaption}
 
\usepackage{enumitem}
\usepackage{engord}
\usepackage{fmtcount}
\usepackage{microtype}
\usepackage{balance}
\usepackage[super]{nth}

%% file: macros.tex
\definecolor{FanColor}{rgb}{0.8,0,0.8}
\newcommand{\fan}[1]{{\color{FanColor}[Fan: #1]}}
\definecolor{YuxinColor}{rgb}{0.54,0.18,0.88}

\newcommand{\warning}[1]{{\it\color{red} #1}}
\newcommand{\toremove}[1]{{\it\color{red} (To remove) #1}}
\newcommand{\note}[1]{{\it\color{blue} #1}}
\newcommand{\nothing}[1]{}

\definecolor{figred}{rgb}{1,0,0}
\definecolor{figgreen}{rgb}{0,0.6,0}
\definecolor{figblue}{rgb}{0,0,1}
\definecolor{figpink}{rgb}{1,0.63,0.63}

\ifthenelse{\equal{\final}{1}}
{

\renewcommand{\fan}[1]{}
\renewcommand{\warning}[1]{}
\renewcommand{\toremove}[1]{}
\renewcommand{\note}[1]{}
}
{}

\renewcommand{\paragraph}[1]{\textbf{#1}}

%% file: Sections/0_abstract.tex
\begin{abstract}

Personalized generation paradigms empower designers to customize visual intellectual properties with the help of textual descriptions by tuning or adapting pre-trained text-to-image models on a few images.
Recent works explore approaches for concurrently customizing both content and detailed visual style appearance. 
However, these existing approaches often generate images where the content and style are entangled.
In this study, we reconsider the customization of content and style concepts from the perspective of parameter space construction.
Unlike existing methods that utilize a shared parameter space for content and style, we propose a learning framework that separates the parameter space to facilitate individual learning of content and style, thereby enabling disentangled content and style.
To achieve this goal, we introduce ``partly learnable projection'' (\textbf{PLP}) matrices to separate the original adapters into divided sub-parameter spaces.
We propose ``\textbf{break-for-make}'' customization learning pipeline based on PLP, which is simple yet effective.
We \textbf{break} the original adapters into ``up projection'' and ``down projection'', train content and style PLPs individually with the guidance of corresponding textual prompts in the separate adapters, and maintain generalization by employing a multi-correspondence projection learning strategy.
Based on the adapters broken apart for separate training content and style, we then \textbf{make} the entity parameter space by reconstructing the content and style PLPs matrices, followed by fine-tuning the combined adapter to generate the target object with the desired appearance. 
Experiments on various styles, including textures, materials, and artistic style, show that our method outperforms state-of-the-art single/multiple concept learning pipelines in terms of content-style-prompt alignment.
\end{abstract}

%% file: Sections/1_intro.tex
\section{Introduction}
Text-to-image (T2I) models based on diffusion technology~\cite{ho2020denoising, song2020denoising, ho2022classifier} have demonstrated remarkable proficiency in generating high-quality images, expanding the imaginative capabilities of humans through textual descriptions.
Represented by Stable Diffusion~\cite{rombach2022high} and Midjourney~\cite{midjourney2023midjourney}, various diffusion models and platforms have been widely applied in the field of creativity design or digital content generation.
Despite the outstanding generalization ability of T2I models, it is challenging for users to generate specific visual concepts using only textual descriptions.

Customized generation approaches have thus been proposed for subject-driven generation by techniques such as tuning the base model with regularization~\cite{ruiz2023dreambooth}, learning additional parameters as pseudo words~\cite{gal2022image, voynov2023p+, alaluf2023neural} or low-rank adaptations~\cite{hu2021lora}.
Most of these approaches, however, only support generating images depicting a single concept (e.g., objects, textures, materials, art style, etc.), leaving the customized generation of multi-concept (e.g. specific content with a specific style) a challenging task.
For example, designers may wish to render specific objects with different textures or materials to examine various effects. Similarly, artists may want to render specific objects in their own distinctive styles.

Multi-concept generation approaches~\cite{kumari2023multi,avrahami2023break} are first proposed to learn and generate different contents by manipulating or constraining cross-attentions mechanisms.  
Different objects would be distinguished on the cross-attention maps using corresponding textual descriptions~\cite{hertz2022prompt}. 
However, the intricate nature of visual style, which is often entangled with content, poses challenges in effectively decoupling content and style concepts due to the shared parameter space and the lack of disentanglement strategies employed by these methods.
As a result, previous approaches cannot be well applied to jointly learn content and style concepts.

To address the content-style customization problem, \citet{zhang2023prospect} recently propose a step-wise pseudo words generation pipeline, which supports combining content and style concepts.
However, relying on step-wise diffusion priors limits ProSpect's generability across different types of visual styles.
More intuitively, ZipLoRA~\cite{shah2023ziplora} merges two independently fine-tuned content and style adaptations using a loss function based on cosine similarity to alleviate the entanglement between content and style.
Nevertheless, the merging process often leads to interference between the parameters of different adapters~\cite{Ortiz-Jimenez_Favero_Frossard_2023}.
This oversight in failing to optimally align the integrated parameters can result in a notable performance degradation of the merged model, leading to ineffective preservation of the distinct qualities of both content and style~\cite{yadav2023resolving}.
Therefore, a method that decouples the learning of content and style, and recombines them in the generation process without interference, is necessary.

In this work, we introduce a two-stage learning approach for customized content-style generation, namely ``break-for-make''.
In the first stage, we propose ``partly learnable projection'' (\textbf{PLP}) matrices to train content and style in separated sub-parameter spaces of low-rank adapters.
Specifically, we freeze certain parameters in both the ``up projection'' and ``down projection'' matrices, allowing separate training of content and style within their respective trainable parameter subsets.
To avoid interference between content and style after matrix multiplication by frozen parameters, we initialize the frozen rows and columns within the projection matrices to approximate orthogonal bases.
To maintain the generalization of the learned content/style PLPs, we utilize a ``multi-correspondence projection'' (\textbf{MCP}) learning strategy to learn unbiased content and style parameter spaces. Specifically, we train customized content in ``up projection'' matrices with diverse reference styles in ``down projection'' matrices and vice versa. 
This approach avoids one-to-one binding between content and style, thereby mitigating the overfitting of content/style PLPs when composing with other corresponding PLPs.
In the second stage, we reconstruct the unified parameter space using the content and style PLP matrices trained in the first stage, then fine-tune the combined adapter to obtain content-style customized results. 
As the specific content and style are learned separately and in a generalized manner during the first stage, fine-tuning (approximately a few dozen steps) is required for the combined adapter to generate images that better align with the content and style references, as shown in Fig.~\ref{fig:head}(a) and (b).

Our contributions can be summarized as follows:
\begin{itemize}
\item We separate the parameter space of low-rank adapters for disentangling the content and style representations and introduce a content-style customization learning pipeline.
\item We propose a Partly Learnable Projection (PLP) with an orthogonal frozen parameters strategy that enables the disentanglement of content and style. During training, a Multi-Corresp-ondence Projection (MCP) mechanism is proposed to maintain generalization.
\item Extensive qualitative and quantitative experiments validate the superior effectiveness of our approach over current baseline methods, particularly in the realms of content and style disentanglement and the preservation of content-style fidelity. 
\end{itemize}

%% file: Sections/2_relate.tex
\section{Related Work}
\subsection{Text-to-Image Customization}
Diffusion models~\cite{ho2020denoising} have demonstrated the capability to produce high-quality images in text-to-image generation~\cite{rombach2022high, saharia2022photorealistic, betker2023improving, chang2023muse}.
Text-to-image customization aims to inject specific concepts or styles into diffusion models to generate diverse images, including different views, poses, scenes, and more~\cite{gal2022image, ruiz2023dreambooth, chen2023subject, gal2023encoder, huang2024creativesynth, wei2023elite, zhang2023inversion}. 
To achieve this, numerous approaches have been proposed across various aspects.
Textual Inversion~\cite{gal2022image} employs inherent parameter space to describe specific concepts and inverts training images back to text embeddings. 
DreamBooth~\cite{ruiz2023dreambooth} fine-tunes backbone models with specific token-images pairs and a prior preservation loss.
Custom diffusion~\cite{kumari2023multi} optimizes a few diffusion model parameters to represent new concepts/styles while enabling fast tuning for multiple concepts jointly.
LoRA~\cite{hu2021lora}, a parameter-efficient fine-tuning approach first revealed for large language models, has proven effective for customization by adapting only a few adaptation parameters. LoRA's lightweight nature and ability to generate customized content/style without full model fine-tuning make it highly flexible. Various LoRA-based methods have been proposed for more effective and efficient training~\cite{hyeon2021fedpara, edalati2022krona, valipour2022dylora, zhang2023adaptive, dettmers2023qlora}.
\citet{po2023orthogonal} design multiple LoRAs to separately train different content and generate multiple contents simultaneously in one image.
By integrating adapter modules, AdapterFusion~\cite{pfeiffer2020adapterfusion} allows adaptation to downstream tasks via fine-tuning only the adapter parameters.
\citet{liu2023cones} propose Cones, a layout guidance method for controlling multiple customized subject generation.
Perfusion~\cite{tewel2023key} introduces a new mechanism locking new concepts' cross-attention Keys to their superordinate category to avoid overfitting, and a gated rank-1 approach to control a learned concept's influence during inference and combine multiple concepts.
NeTI~\cite{alaluf2023neural} and ProSpect~\cite{zhang2023prospect} introduce an expanded text-conditioning space over diffusion time steps for fine-grained control.
These concept-customized generation methods primarily focus on the quality of generated outputs, addressing general concept customization. In contrast, we focus mainly on the fusion generation of customized content and style.

\input{Figures/pipeline}

\subsection{Customized Content Style Fusion}
The goal of the content-style customization is to generate an image that incorporates specific content and style based on reference images, while ensuring the unique characteristics of both content and style are distinctively represented and aligned with prompts.
Previous works jointly train content and style on customized generation models~\cite{gal2022image, ruiz2023dreambooth, kumari2023multi}. During inference, these methods generate images blending both content and style based on given prompts. 
However, these straightforward approaches do not optimize the learning between content and style, often resulting in their entanglement in the generated results.
DreamArtist~\cite{dong2022dreamartist} employs a positive-negative prompt-tuning learning strategy for customized generation and discusses content-style image fusion in the experiments.
SVDiff~\cite{han2023svdiff} fine-tunes the singular values of weight matrices and proposes a Cut-Mix-Unmix data-augmentation technique to help multi-subject and content-style image generation.
StyleDrop~\cite{sohn2023styledrop} improves the quality of generating stylized images via iterative training with human or automated feedback.
ProSpect~\cite{zhang2023prospect} leverages learning word embeddings specific to content and style, incorporating them at different diffusion time steps to control customized content-style image generation. 
However, relying on step-wise diffusion priors limits ProSpect's generability across different content and visual styles.
Recent work ZipLoRA~\cite{shah2023ziplora}learns hybrid coefficients to optimize conflicts arising when merging two separately trained LoRAs, partially mitigating disentanglement issues. However, it concurrently modifies the distribution of learned parameters, subsequently influencing reconstruction outcomes.
Compared to related approaches, our proposed ``partly learnable projection'' and ``multi-correspondence projection learning'' strategy trains content and style separately in different sub-parameter spaces within low-rank adaptations with data augmentation.
This effectively disentangles content and style in generated images while maintaining high image fidelity.

%% file: Figures/pipeline.tex
\begin{figure*}[ht!]
  \centering
  \includegraphics[width=1\linewidth]{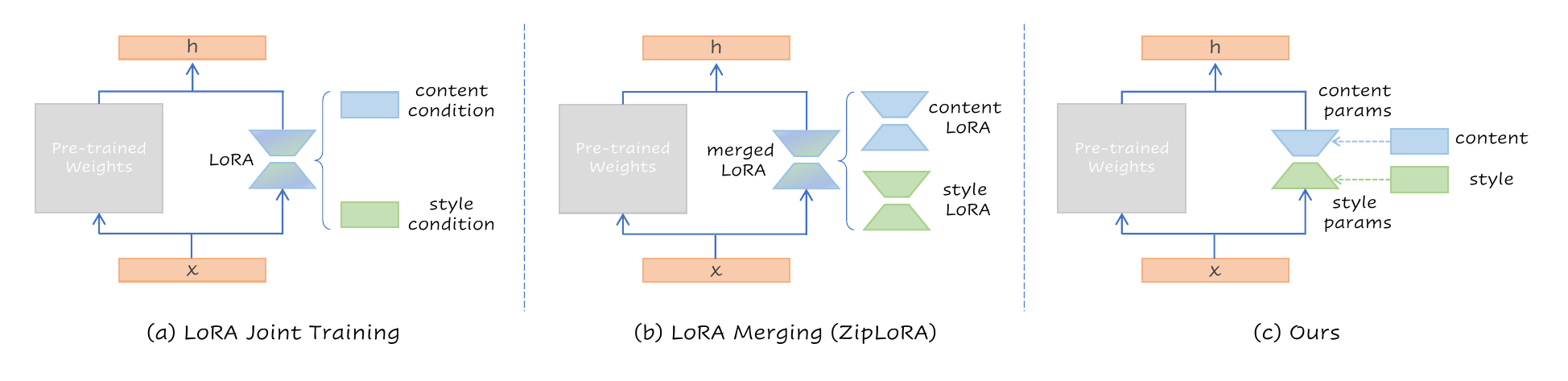}
  \caption{\textbf{Frameworks of the two main approaches and ours for customized content-style image generation.} LoRA joint training incorporates image-text pairs to fine-tuning the overall model parameters. ZipLoRA \cite{shah2023ziplora} effectively merging independently trained content and style LoRAs, then add to per-trained weights to generaet images of the customized content and style. Our method trains content and style in separated parameter subspaces of LoRA, results in disentanglement of content and style while maintaining high level of fidelity.}
  \label{fig:pipeline}
\end{figure*}

%% file: Sections/3_method.tex
\section{vanilla solutions for Content-Style Customization}
\label{sec:vanilla}
In this section, we first introduce the task definition of content-style customization in image generation. 
Then, we review existing related methods, including the basic low-rank adaptation fine-tuning method, joint training method, and merging after independent training method. Note that our primary focus is on methods based on low-rank adaptations, as these are both efficient and effective for fine-tuning large T2I models. We then investigate why these methods fail to generate images of disentangled content and style faithfully. In response to these challenges, we propose our novel solution.

The goal of content-style customization is to generate images that effectively present both user-specified content and style while ensuring their unique characteristics are distinctively represented~\cite{zhang2023prospect,shah2023ziplora}.
Formally, given a content reference image $I_{c}$, a style reference image $I_{s}$, and a prompt $P$, we aim to generate an output image $I_{out}$ that contains the same content as $I_{c}$, and has the same style as $I_{s}$, while aligning with the provided prompt $P$.

However, accurately generating specific content while effectively rendering it in a reference style without conflict is challenging. 
For example, the inherent style of the user-provided content image may interfere with the reference style, leading to style conflict when generating a customized content-style image. 
LoRA~\cite{hu2021lora}, a lightweight adaptation method, has been applied in customized generation, enabling effective learning of content and style within the LoRA module.
This motivates us to train content and style in separate subspaces of the low-rank adaptation parameters.

\paragraph{Low-Rank Adaptation Fine-Tuning.}
LoRA~\cite{hu2021lora} is an efficient adaptation strategy for fine-tuning large pre-trained models while retaining high quality.
Initially proposed for fine-tuning large language models, LoRA has also proven suitable for fine-tuning vision models like diffusion models for text-to-image generation.
For a per-trained weight matrix $W_{{o}} \in \mathbb{R}^{m \times n}$, each LoRA module consists of an up-projection matrix $W_{up} \in \mathbb{R}^{m \times r}$ and a down-projection matrix $W_{{down}} \in \mathbb{R}^{r \times n}$, where the rank $r \ll min(m,n)$. Given an input $z$, during training, the forwards pass is:
\begin{equation}
I = W_{0}z + W_{{up}}W_{{down}}z,
\end{equation}
and only $W_{{up}}$ and $W_{{down}}$ are updated to find a suitable adaptation $\Delta W = W_{{up}}W_{{down}}$.
In this work, we incorporate LoRA modules into the cross-attention components of the diffusion model for fine-tuning \cite{cloneofsimo2023lora}.
After that, we can directly merge the LoRA module with the per-trained weight matrix and obtain new weights $W = W_{0}+\Delta W$, which can perform inference as usual.

\paragraph{Jointing Training.}
A straightforward method for customized content-style generation is jointly training LoRA modules with customized content images and style images.
In simple terms, LoRA modules $W$ for learning specific content and style are trained using a squared error loss function as follows:
\begin{equation}
    L = [\lVert \hat{W}_{\theta}(z_{c}|c_{c},t) - x_{c} \rVert ^{2}_{2}] + [\lVert \hat{W}_{\theta}(z_{s}|c_{s},t) - x_{s} \rVert ^{2}_{2}],
    \label{eq:loss_joint}
\end{equation}
where $(z_{c}, c_{c}, x_{c})$ and $(z_{s}, c_{s}, x_{s})$ are data-conditioning-target pairs of the specific content and style (image latent, text embeddings and target images), respectively. $t$ is diffusion process time $t \sim ([0,1])$, and $\theta$ represents model parameters.
The training pipeline is presented in Fig.~\ref{fig:pipeline} (a).
However, this training approach mixes the parameter spaces of content and style during the training stage, resulting in the entanglement of content and style when weights $W$ multiplied with the input, as analyzed in Fig.~\ref{fig:fig_3_analysis} (a).

\paragraph{Merging after Independent Training.}
Another primary method involves independently training two LoRA modules — one dedicated to content and the other to style — in the first stage. Subsequently, in the second stage, these modules are merged with certain constraints, as shown in Fig.~\ref{fig:pipeline}(b).
Given a set of learned LoRA weights $\Delta W_{i}$ optimized on content and style, the merged weight is simply given by
\begin{equation}
W_{merged} = W_{0} + \sum_{i}\lambda_{i} W_{i},
\end{equation}
where $\lambda_{i}$ is a scalar representing the relative strength of content and style.
However, directly merging two independently trained LoRAs may lead to parameter conflict. Specifically, when merging a parameter that is influential for one LoRA but redundant for the other, the influential value may be obscured by the redundant values, resulting in a decrease in overall effectiveness~\cite{yadav2023resolving}.
This interference leads to the loss of content and style features learned during independent training stages, as analyzed in Fig.~\ref{fig:fig_3_analysis} (b).
ZipLoRA~\cite{shah2023ziplora} learns mixing coefficients for both content and style LoRAs to mitigate conflicts. Nevertheless, to some extent, it affects the distribution of content and style parameters learned during the training phase. Although this approach shows improved disentanglement performance, the fidelity of reconstruction is somewhat reduced. This motivates us to pursue separate training for content and style, aiming to achieve both precise reconstruction on customized content and style as well as effective disentanglement.

\input{Figures/fig_3_analysis}

\section{Our method}
\label{sec:method}
In this section, we first introduce our proposed ``partly learnable projection'' (\textbf{PLP}) method, a parameter separation training framework for LoRA that enables better control over the training parameters.
This facilitates the generation of images that are more faithfully aligned with the specified conditions while maintaining higher fidelity.
We then present ``Multi-Correspondence Projection Learning'' (\textbf{MCP}), a technique for training content and style representations during the customization process to mitigate overfitting between the two.
By utilizing both the proposed \textbf{PLP} and \textbf{MCP} methods, we enable the generation of customized content-style images that achieve effective disentanglement of content and style, while also preserving a high degree of image fidelity.

\subsection{Partly Learnable Projection}
To address the aforementioned issues, we propose ``Partly Learnable Projection'' (\textbf{PLP}) matrices to separate the LoRA module and search for the optimal content and style parameters within distinct sub-parameter spaces.
Specifically, we consider a LoRA module $\Delta W$ with input dimension $n$, rank $r$, and output dimension $m$. The $W_{down}$ and $W_{up}$ matrices of $\Delta W$ are decomposed into two submatrices along the feature dimension, respectively. 
The $W_{up}$ can be formed as: 
\begin{equation}
W_{up} = \begin{bmatrix}
A & B
\end{bmatrix}^{-1}, 
\end{equation}
where
\begin{equation}
A = \begin{bmatrix}
A_{11}  & \cdots & A_{1r} \\
\vdots  & \ddots & \vdots \\
A_{d1}  & \cdots & A_{dr}
\end{bmatrix},  B = \begin{bmatrix}
B_{(m-d)1} & \cdots & B_{(m-d)r} \\
\vdots & \ddots & \vdots \\
B_{m1} & \cdots & B_{mr}
\end{bmatrix}.
\label{eq:submatrixA&B}
\end{equation}
Similarly, the $W_{down}$ matrix can be formed as:
\begin{equation}
W_{down} = \begin{bmatrix}
C & D
\end{bmatrix}, 
\end{equation}
where
\begin{equation}
C = \begin{bmatrix}
C_{11}  & \cdots & C_{1d} \\
\vdots  & \ddots & \vdots \\
C_{r1}  & \cdots & C_{rd}
\end{bmatrix},  D = \begin{bmatrix}
D_{1(n-d)}  & \cdots & D_{1n} \\
\vdots  & \ddots & \vdots \\
D_{r(n-d)}  & \cdots & D_{rn}
\end{bmatrix}.
\label{eq:submatrixC&D}
\end{equation}
According to the rules of partitioned matrix multiplication, we have 
\begin{align}
   \Delta{W} &= W_{up}W_{down} \\
                  &= \begin{bmatrix}
                        W_{ul}&W_{ur}\\
                        W_{dl}&W_{dr}
                    \end{bmatrix},
    \end{align}
\label{eq:delta_w}
where
\begin{equation}
W^{ul}_{i,j} = \sum_{r}A_{i,r}C_{r,j},
\label{eq:W_ul}
\end{equation}
\begin{equation}
W^{ur}_{i,j} = \sum_{r}A_{i,r}D_{r,j},
\label{eq:W_ur}
\end{equation}
\begin{equation}
W^{dl}_{i,j} = \sum_{r}B_{i,r}C_{r,j},
\label{eq:W_dl}
\end{equation}
\begin{equation}
W^{dr}_{i,j} = \sum_{r}B_{i,r}D_{r,j}.
\label{eq:W_dr}
\end{equation}
Here, $d$ represents the feature dimension of the fixed parameters. Adjusting the size of $d$ implies modifying the ratio of frozen to trainable parameters within the matrix. We will discuss it in Section~\ref{sec:ablation}. 
\input{Figures/Multi-correspond}
After multiplication, we obtain a partitioned matrix, which can be visualized as the original matrix decomposed into a set of horizontal and vertical submatrices.

We propose PLP with orthogonal parameters for better disentanglement of content and style during training. 
Specifically, the matrices $A$ and $C$ in Eq.~(\ref{eq:submatrixA&B}) and Eq.~(\ref{eq:submatrixC&D}) are kept frozen during the training process.
We initialize $A$ and $C$ as approximately orthogonal to reduce redundant parameters and achieve better disentanglement of content and style:

\begin{align}
    W^{ul}_{i,j} &= \sum_{r}A_{i,r}C_{r,j}, \\
                    &=0.
    \end{align}
\label{eq:W_ul_ij}

We can notice that, the up-right part of $\Delta{W}$ in Eq.~(\ref{eq:W_ur}) represents \textbf{only} the parameters of submatrix $D$ only.
Similarly, the down-left part of $\Delta{W}$ in Eq.~(\ref{eq:W_dl}) represents \textbf{only} the parameters of submatrix $B$ only, 
$W^{dr}_{i,j}$ in Eq.~(\ref{eq:W_dr}) relates to $B$ and $D$, allowing us to learn the interactive features between them.

The forward pass during training yields:
\begin{equation}
    I = W_{0}z + \begin{bmatrix}
    0&\sum_{r}A_{i,r}D_{r,j} \\
    \sum_{r}B_{i,r}C_{r,j} & \sum_{r}B_{i,r}D_{r,j}
    \end{bmatrix}z,
\end{equation}
where $A_{i,r}$ and $C_{r,j}$ are frozen during training.

So far, we have demonstrated that our proposed method based on partitioned matrices can effectively separate the parameters associated with content and style. This enables the multiplication of input features with the corresponding parameters during training, thereby distinctly representing the acquired content and style in different subspaces of the parameter space. Consequently, this mitigates the entanglement between content and style during image generation while preserving a high degree of fidelity. 

Our method is illustrated in Fig.~\ref{fig:fig_3_analysis} (c), which indicates that after separating the LoRA module into two parts and performing forward matrix multiplication, the resulting partitioned matrices exhibit the following advantageous characteristics: the top-left part consists of zeros due to the multiplication of orthogonal vectors, the top-right part represents the style submatrix for learning style image feature parameters, the bottom-left part represents the content submatrix for learning content image feature parameters, and the bottom-right part is utilized for learning interactive feature parameters between content and style.
The four distinct parts in the partitioned matrix demonstrate that our method successfully separates content and style for training in different LoRA parameter subspaces. This circumvents the parameter conflict issues introduced by merging methods and allows us to obtain disentangled content and style feature representations. Meanwhile, the interactive parameters between content and style enable the generation of more naturalistic fusion images with high visual quality.

\subsection{Multi-Correspondence Projection Learning}
When training specific content and style in a one-to-one manner, overfitting issues may arise, resulting in suboptimal performance when reconstructing the content-style modules in the second stage for image generation.
To mitigate this problem between content and style during training, we introduce a multi-correspondence projection (``\textbf{MCP}'') learning method involving diversified content-style training data pairs. Specifically, when training for a particular content, we update the parameters of $B$ in Eq.~(\ref{eq:submatrixA&B}) with the particular content image and update the parameters of $D$ in Eq.~(\ref{eq:submatrixC&D}) with various style images, vice versa.
In simple terms, a LoRA model $W$ for learning specific content is trained using a squared error loss function as follows:
\begin{equation}
    L = [\lVert \hat{W}_{\theta}(z_{c}|c_{c},t) - x_{c} \rVert ^{2}_{2}] + \frac 1n \cdot\sum_{i=1}^{n}[\lVert \hat{W}_{\theta}(z_{s}|c_{s},t) - x_{s} \rVert ^{2}_{2}],
    \label{eq:loss}
\end{equation}
where $(z_{c}, c_{c}, x_{c})$ and $(z_{s}, c_{s}, x_{s})$ are data-conditioning-target triplets of the specific content and diverse styles (image latents, text embeddings, and target images), respectively. $n$ represents the number of different styles. $t$ is the diffusion process time $t \sim ([0,1])$, and $\theta$ represents the model parameters.
The loss function for training the style LoRA model is similar to Eq.~(\ref{eq:loss}).
This training approach prevents overfitting issues that arise when learning specific content-style pairs (see the ablation study in Fig.~\ref{fig:ablate_compare}), simultaneously enhancing the method's generalization ability and improving the effectiveness of diverse content-style combinations.

\paragraph{Inference.} After training, we obtain $LoRA_{c}$ which contains the learned parameters of the specific content, and $LoRA_{s}$ which contains the learned parameters of the specific style. We then combine the up-projection part of $LoRA_{c}$ with the down-projection part of $LoRA_{s}$ to reconstruct $LoRA_{f}$ as the fusion adapters. With a few dozen fine-tuning steps of $LoRA_{f}$ on the given content and style images, we can effectively obtain the final adapters capable of generating content-style disentangled images with high fidelity. For single content or style generation, we can directly perform inference using the learned $LoRA_{c}$ or $LoRA_{s}$, respectively.

\input{Figures/fig_compare_multistyle}

%% file: Figures/fig_3_analysis.tex
\begin{figure}[t!]
  \centering
  \includegraphics[width=1\linewidth]{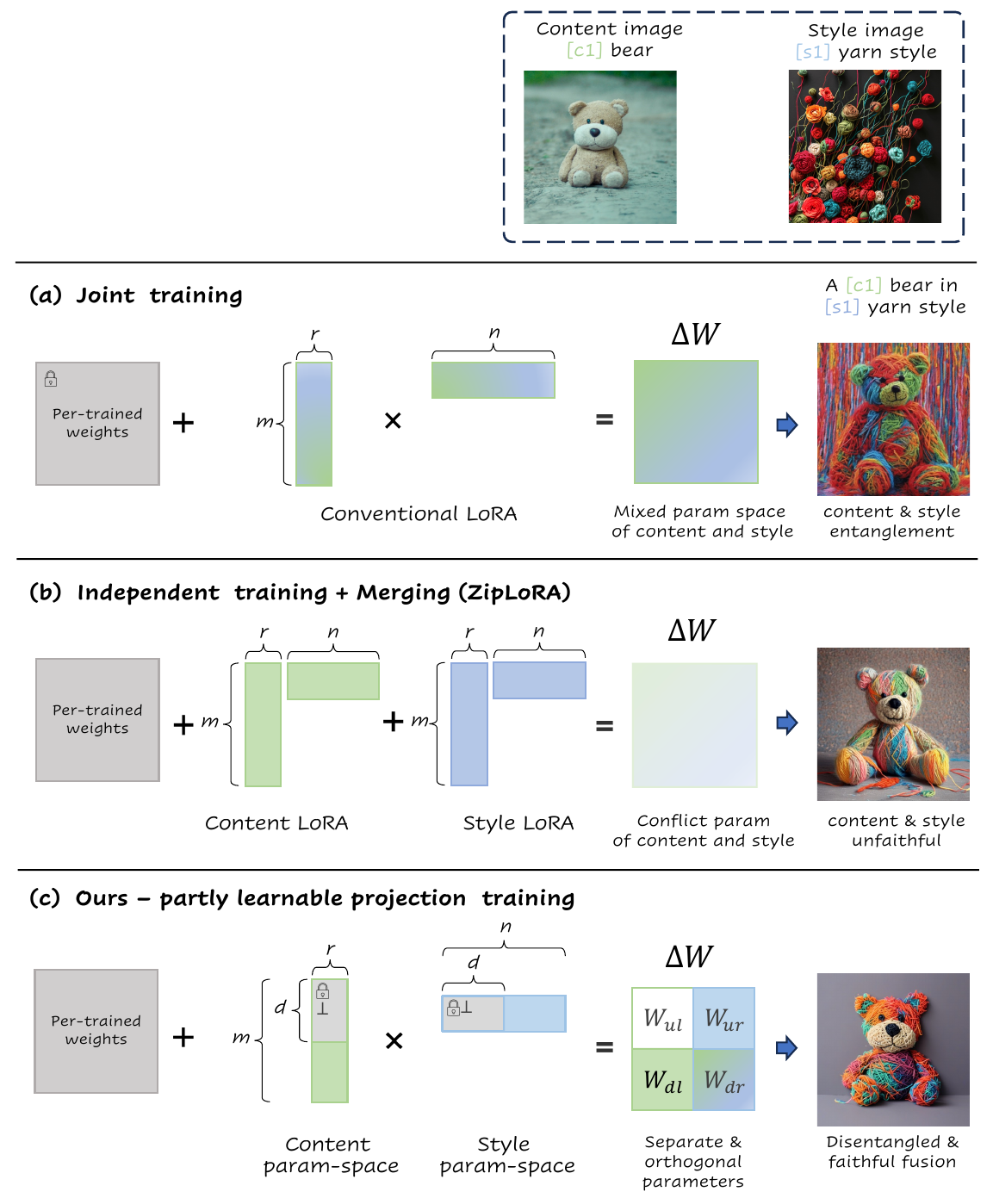}
  \caption{\textbf{Analysis of the two main methods and ours using LoRA for customized content-style image generation.} Joint training LoRA will mix the parameter space of content and style, leads to entanglement of content and style. Merging LoRAs after independent training has problem of conflict parameters from content LoRA and style LoRA, leads to content and/or style unfaithful after fusion. Our proposed method trains the content and style in separate parameter subspaces of the LoRA modules, with orthogonal fixed parameter spaces, resulting in disentangled and faithful fusion of content and style.}
  \label{fig:fig_3_analysis}
\end{figure}

%% file: Figures/Multi-correspond.tex
\begin{figure}[!t]
  \centering
  \includegraphics[width=1\linewidth]{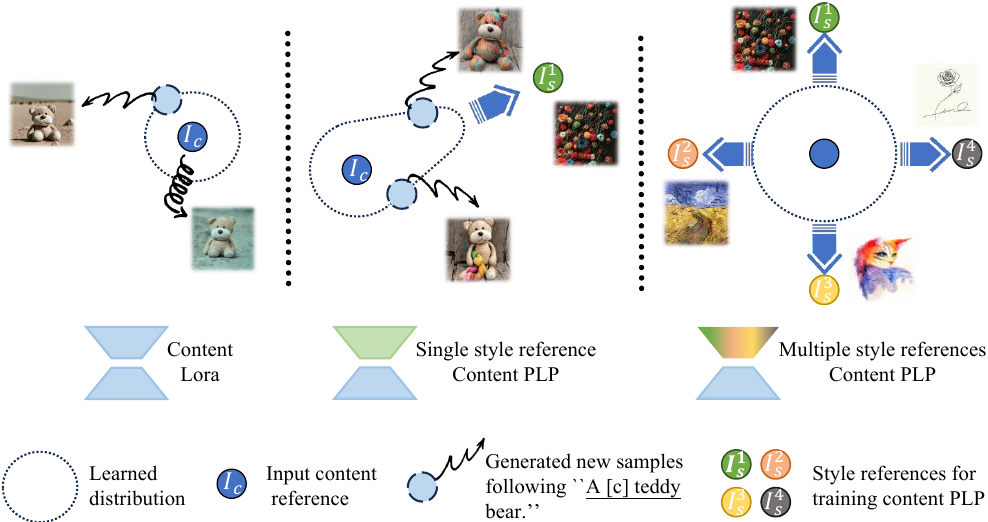}
  \caption{\textbf{Illustration of the multi-correspondence projection.}
  We present the learned content distribution on the left of the top row.
  When training specific content and style in a one-to-one manner, the content will tend to overfit to the specific style, as illustrated on the middle of the top row.
  By leveraging our proposed multi-correspondence projection, we learn multiple styles with the content in PLP, enhance the generalization of the learned content.}
  \label{fig:Multi-correspond}
\end{figure}

%% file: Figures/fig_compare_multistyle.tex
\begin{figure*}[ht!]
  \centering
  \includegraphics[width=1\linewidth]{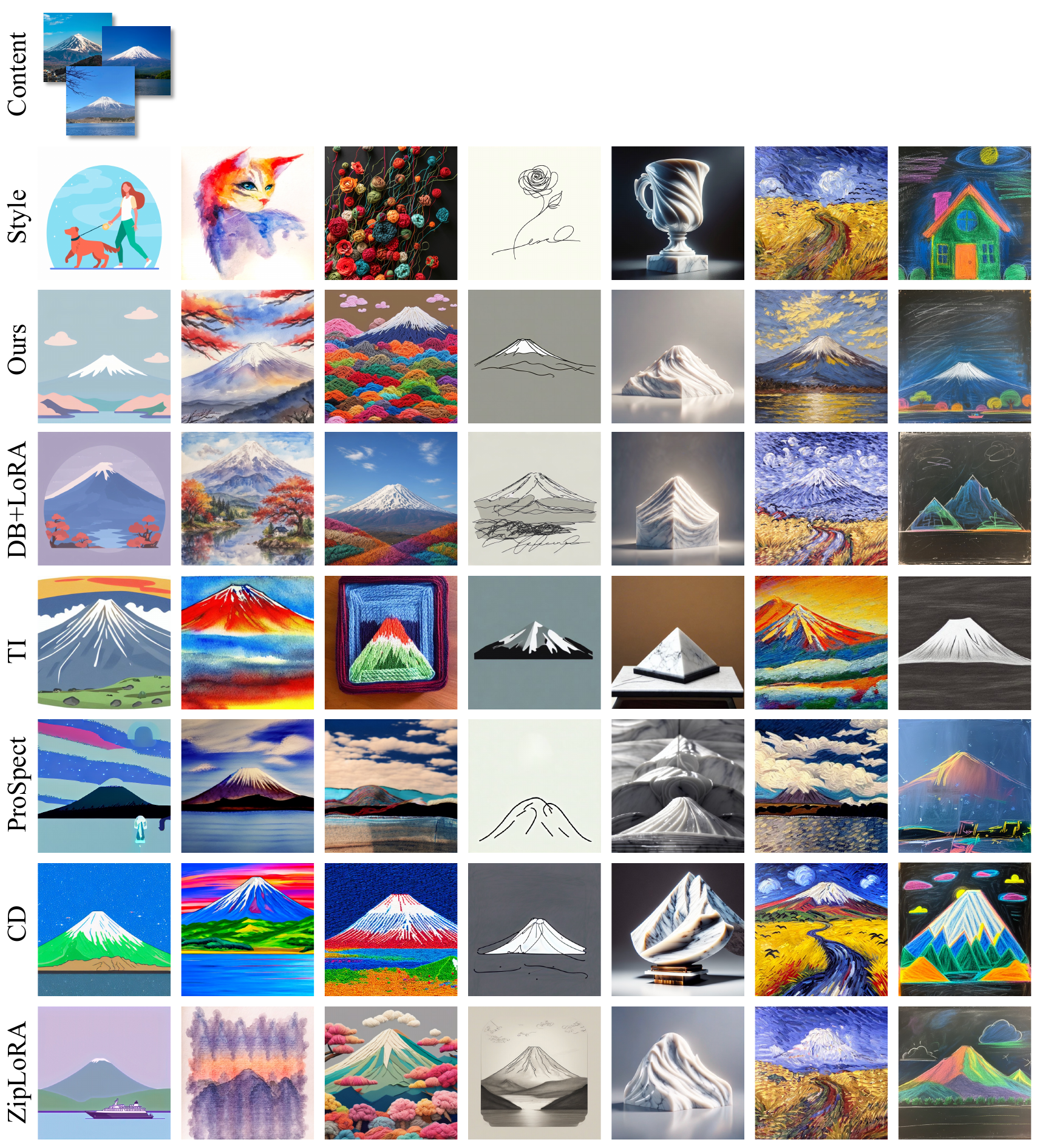}
  \caption{\textbf{Qualitative evaluation and comparison of DB+LoRA, TI, ProSpect, CD, ZipLoRA and our method in diverse styles}. We present the results of customized generation of the same content and different styles. Results indicate that our method generates harmonious fusion images of the content and the style, while preserving the disentanglement of content and style, as well as maintaining high-level fidelity of them.}
  \label{fig:fig_compare_multistyle}
\end{figure*}

%% file: Sections/4_experiments.tex
\input{Figures/fig_main_compare}

\section{Experiments}
In this section, we conduct qualitative comparisons and quantitative evaluations to demonstrate our method outperforms state-of-the-art customized content-style fusion baselines.
We also conduct ablation studies to analyze the impact of certain crucial modules in our approach on the generation results.

\textbf{Datasets.} For fair and unbiased evaluation, we use concept images and style images from related works \cite{shah2023ziplora,zhang2023prospect,ruiz2023dreambooth, gal2022image} together with diverse images from the Internet. For training content images, we collect three to five images of the same content and five different styles, each style consisting of one image. For training style images, we collect one to three images of the same style and five different contents, each content consisting of one image.

\textbf{Compared Methods.} We compare our method against state-of-the-art baselines on the task of content-style customization.
\begin{itemize}
    \item Dreambooth+LoRA (DB+LoRA)~\cite{cloneofsimo2023lora} leverages LoRA for customized content and style generation. We jointly train the LoRA with content and style images and their corresponding prompts. In each epoch, we first update parameters based on content loss for the same model, followed by updating parameters based on style loss.
    \item Textual inversion (TI)~\cite{gal2022image} inverts a customized image back to text embeddings and bands it with a token. Then, the token can be composed into a prompt to generate related content or style in the output image. We learn content and style on two separate tokens and simultaneously incorporate these two tokens into the prompt for content-style customization.
    \item ProSpect \cite{zhang2023prospect} proposes a novel approach that adds different conditions to the diffusion model in different generation steps to achieve fine-grained and controllable generation. In our experiment, we add customized content and style as conditions in different steps of generation to achieve content-style fusion, according to their paper.
    \item Custom Diffusion (CD) \cite{kumari2023multi} proposes an efficient fine-tuning method for simultaneously generating multiple customized content or content with style. We followed the official open-source code, conducting joint training for content and style.
    \item ZipLoRA \cite{shah2023ziplora} is a recently released method that provides a novel approach to merge trained content and style LoRAs by learning mixing coefficients for LoRAs. As official codes of ZipLoRA have not been released yet, we evaluate this method with a popular implementation in the GitHub community~\cite{mkshing2023ziplora}. We initially train the content and style models separately and then perform LoRA merging based on the parameters specified in the paper.
\end{itemize}

\paragraph{Metrics.}
We primarily conduct qualitative and quantitative comparisons between our method and baseline methods. For qualitative comparisons, we primarily present and compare the visual quality of the generated images. For quantitative comparisons, we mainly assess three metrics: content alignment and style alignment between the generated images and reference images, as well as text alignment between the generated images and the corresponding prompts. 
Following quantitative experiment settings of ProSpect~\cite{zhang2023prospect} and ZipLoRA~\cite{shah2023ziplora}, we compare cosine similarities between CLIP~\cite{ilharco_gabriel_2021_5143773} (for style and prompt) and DINOv2~\cite{oquab2023dinov2} features (for content) of the generated images and reference contents, styles and prompts respectively.

\input{Figures/first_fig}

\paragraph{Implementation Details.} 
In our experiments, we utilized Stable Diffusion XL v1.0~\cite{podell2023sdxl} with default hyperparameters and set a base learning rate of $0.0001$. 
During training, we set the batch size to $1$, text encoders of SDXL are kept frozen, and the refiner of SDXL is not utilized. 
Based on the orthogonal fixed parameters we proposed, we train LoRA modules with the same size as the input and output feature dimensions. 
The rank of LoRA is set to $64$. 

\input{Tables/alignment.tex}

\subsection{Main Results}
In this section, we present Qualitative and Quantitative Comparisons between our method and baseline approaches. Additionally, we showcase more of our results with diverse contents and styles in Fig.~\ref{fig:first_fig}.

\paragraph{Qualitative Comparison.}
We compare our method with five content-style customization methods: DreamBooth+LoRA (DB+LoR-A), Textual Inversion (TI), ProSpect, Custom Diffusion (CD) and ZipLoRA.
We first present the results of generating the same content image with multiple style images in Fig.~\ref{fig:fig_compare_multistyle}, then we present the same style image with multiple content images in Fig.~\ref{fig:fig_main_compare}.
Results indicate that our methods successfully disentangle content and style in one image while maintaining a high level of fidelity.
The DB+LoRA method usually generates images of unnatural content style fusion (the result of ``mountain'' with ``yarn style'' and ``oil painting style'' in Fig.~\ref{fig:fig_compare_multistyle}) and images of the mixed style (``vase'' and ``teapot'' with ``glass style'' in Fig.~\ref{fig:fig_main_compare}), the observed entanglement phenomenon aligns with the analysis presented in Section~\ref{sec:vanilla}.
The TI method only updates parameters in the text embedding space, thus having a relatively weaker learning capability. At times, it struggles to accurately learn content/style features, leading to a decrease in the fidelity of generated images (``mountain'' with ``Minimalism painting style'' and ``marble style'' in  Fig.~\ref{fig:fig_compare_multistyle}, the loss of feature ``transparent glass'' in ``glass style'' in Fig.~\ref{fig:fig_main_compare}).
ProSpect learns the reference image with a specific token and trains the embedding of this token, then adds it as a condition in different steps during inference. This method has achieved effective control over content and style to some extent, as seen in examples such as ``vase'' and ``teapot'' in Fig.~\ref{fig:fig_main_compare}, shape and material are presented in generated images. However, it is constrained by its learning capability, which leads to low-quality content-style customization results (the result of ``mountain'' with ``watercolor painting style'' and ``yarn style'' in Fig.~\ref{fig:fig_compare_multistyle}). 
The CD also encounters entangling issues between content and style. In cases of ``glass style'' with ``vase'' and ``teapot'' in Fig.~\ref{fig:fig_main_compare}, the reference images of content influence the style of the generated images.
In the case of ZipLoRA, the generated results may not accurately present the reference content or style. 
For example, in instances like ``vase'' and ``teapot'' in Fig.~\ref{fig:fig_main_compare}, the outputs of ZipLoRA lack the texture style of ``transparent glass'' in the reference set. 
In instances of generating ``mountain'' with ``oil painting'' style and ``blackborad painting'' style, the mountain cannot be generated faithfully as the reference.
This also reflects the manifestation of fidelity degradation due to parameter conflicts.
Compared with the above methods, our method maintains a high level of fidelity and harmonious content-style interaction when generating various styles for the same content. This also demonstrates the strong generalization capability of our approach.
One more interesting thing is that the instance of the ``sticker style'' images includes a dual style, encompassing both sticker and cartoon styles.
When evaluating it as the reference, our method successfully generates images in the sticker style. It simultaneously transfers the content into a cartoon style, while the results of other methods are kept in a realistic style.

\input{Figures/fig_editability}

\paragraph{Quantitative Comparison.}
We present quantitative comparison results in Table~\ref{tab:alignment}, evaluating the style-alignment, prompt-alignment (using CLIP feature extraction), and content-alignment (using DINO feature extraction) metrics.
Additionally, we report the average of these three metrics, where higher values indicate better performance.
Our method achieves the highest average score among all baselines, suggesting it generates customized content-style images that align well with the content and style references while corresponding to the given prompt.
Note that in the content-alignment metric, our score is not the highest because other methods tend to generate images that retain more features from the content reference images.
However, this could compromise accurate expression of style and adherence to the prompt in the generated images, affecting the effectiveness of style transfer and prompt alignment, as indicated by the lower style-alignment and prompt-alignment metrics for other methods in Table~\ref{tab:alignment}.
Additionally, the comparative display in Fig.\ref{fig:fig_compare_multistyle} and Fig.~\ref{fig:fig_main_compare} supports this observation.

\input{Figures/tsne}

\subsection{Editability Evaluation and Comparison}
We evaluate and compare the editability of our method against other baselines by generating customized content-style fusion images using a diverse set of prompts.
For a fair comparison, the prompts and results of the ZipLoRA are obtained from their original paper. 
As illustrated in Fig.~\ref{fig:fig_editability}, ZipLoRA is generally effective in generating customized content-style images that align well with the provided prompts.
However, in some details, ZipLoRA tends to lose certain characteristics of the reference image, such as the ears and mouth in the ``wearing a hat'' example, and the overall appearance in the ``in a boat'' and ``driving a car'' examples. In contrast, our method maintains better consistency with the reference image in these generated results.
Additionally, we showcased more generation results from diverse prompts in the bottom two rows of Fig.~\ref{fig:fig_editability}.
These results demonstrate high alignment with the prompts while maintaining a high level of disentanglement between content and style, as well as preserving the fidelity of content and style representations.
Overall, our method exhibits superior editability compared to existing baselines, enabling the generation of customized content-style images that faithfully integrate the provided prompts while retaining the desired characteristics of the reference content and style.

\subsection{Visualizing and Comparing Parameter Distributions for Our Method and Baseline Methods}
We employ t-SNE~\cite{van2008visualizing} (t-Distributed Stochastic Neighbor Embedding) to visualize the high-dimensional parameter distributions of the low-rank adapters from our method and the joint training baseline.
Specifically, we use t-SNE to reduce the column dimension of the low-rank adapters parameters to 2 dimensions.
Fig.~\ref{fig:tsne}(a) depicts the parameter distribution of the low-rank adapters from our proposed method after applying t-SNE for dimensionality reduction and visualization.
For a fair comparison, we set the orthogonal part of the joint training baseline's parameters to zero to align with our methods' parameter formulation. Fig.~\ref{fig:tsne}(b) shows the t-SNE visualization of the resulting joint training baseline's low-rank adapter parameter distribution.
Additionally, Fig.~\ref{fig:tsne}(c) presents the parameter distribution of the joint training baseline after t-SNE visualization, but without zeroing out the orthogonal parameter components.
The results indicate that, after training, compared with the joint training baseline method, our method successfully separates the parameter space of the low-rank adapters from a uniform distribution.

\subsection{Concept and Style Reconstruction after First Stage Training}
The proposed two-stage training paradigm employs a multi-corresp-ondence projection learning strategy in the first stage to accurately learn the specific content and style features.
Consequently, in the second stage, only a few dozen iterations of fine-tuning are required to generate customized content-style fusion images. 
Fig.~\ref{fig:fig_reconstruct} illustrates that our approach has accurately learned the features of content or style in the first stage of training and maintained a high level of fidelity for individual content and style after the second stage of fine-tuning.
Moreover, this multi-correspondence projection learning strategy prevents overfitting between content and style, thereby enabling the generation of more diverse results based on prompts.
We also present images generated from directly combined adapters without fine-tuning in Fig.~\ref{fig:fig_reconstruct}, these results verify that content and style are disentangled in the first stage and have better effects after undergoing the fine-tuning process.

\input{Figures/fig_reconstruct}

\subsection{Comparison with Other Two-Stage Content-Style Customization Paradigms}
For the task of customized content-style image generation, we also evaluate other two-stage approaches that involve learning specific content/style in the first stage and subsequently learning or editing style/content~\cite{bau2019semantic, hertz2022prompt, avrahami2023blended, avrahami2023spatext, brooks2023instructpix2pix, mokady2023null, kawar2023imagic, parmar2023zero, balaji2022ediffi} based on the previous results in the second stage. 
In our experiments, we learn the content of reference images in the first stage and learn or edit style in the second stage. We leverage NULL-text Inversion ~\cite{mokady2023null}, a SOTA real image editing method to edit style in the second stage. 
The results are presented in Fig.~\ref{fig:compare_edit}. 
We observe that both the two-stage training and editing methods share similar drawbacks, primarily the entanglement between content and style features.
For instance, when generating the ``glass'' style, the ``teddy bear'' retains plush features, and the ``vase'' and ``teapot'' retain opaque material from the content reference. 
In the case of the ``sticker'' style, these two methods only generate the contours as the ``sticker'' style, while the content of the sticker still reflects the realistic style depicted in the content reference image.
Furthermore, the editing-based approach often necessitates complex prompts to accurately describe the features of the reference image, thereby increasing the difficulty of precisely customizing content-style generation.
In contrast, our method effectively disentangles the content and style of the reference image, blending them together to generate high-quality customized content-style images without the need for complex prompts.
Our approach demonstrates superior performance in achieving faithful content-style fusion compared to both the two-stage training and editing methods.

\input{Figures/compare_edit}
\input{Figures/user_study_abtest_1}
\subsection{User Study}
We conduct a user study to assess the images generated by our method and other baseline methods, employing five comparative methods: DB+LoRA, TI, ProSpect, Custom Diffusion (CD), and ZipLoRA.
A total of $45$ participants took part in the survey, including 20 researchers in computer graphics or computer vision. 
Among the participants, five are aged between 10 and 19 years old, 36 are aged between 20 and 39 years old, and four are aged between 40 and 60 years old. 
Additionally, there are $23$ female participants and 22 male participants.
The evaluation primarily includes: I. Alignment of content between generated images and reference images; II. Alignment of style between generated images and reference images; III. Overall alignment of content and style between generated and reference images; IV. The success rate of generating images with custom content and style; IV. Stability of the generated results.
\begin{itemize}
    \item \emph{User Study I.}
    Alignment of content between generated images and reference images.
    One of the objectives of the content-style customization task is to ensure the alignment of content between the generated image and the reference image. 
    In this user study, participants were tasked with selecting the image that most closely aligned with the given content reference image from six images generated using different methods (including our proposed method and baseline methods).
    Our method received 55.00$\%$'s preference while DB+LoRA, TI, ProSpect, CD, ZipLoRA received 7.78$\%$, 6.48$\%$, 7.41$\%$, 4.44$\%$, 18.89$\%$, respectively.
    Results are presented on the left of Fig.~\ref{fig:user_study_ab_test_1}(a), indicating that our generated images have a higher level of content alignment with the reference images compared to baseline methods.
    \item \emph{User Study II.}
    Alignment of style between generated images and reference images.
    We also need to evaluate the alignment of style between the generated image and the reference image. 
    In this user study, participants were tasked with selecting the image that most closely aligned with the given style reference image from six images generated using different methods (including our proposed method and baseline methods).
    Our method received 79.07$\%$'s preference while DB+LoRA, TI, ProSpect, CD, ZipLoRA received 5.56$\%$, 2.96$\%$, 0.93$\%$, 2.04$\%$, 9.44$\%$, respectively. 
    Results are presented in the middle of Fig.~\ref{fig:user_study_ab_test_1}(a), indicating that our generated images have a higher level of style alignment with the reference images compared to baseline methods.
    \item \emph{User Study III.}
    Overall alignment of content and style between generated images and reference images.
    Participants were asked to provide an overall assessment of the alignment between the generated images and both the content reference images and the style reference images, selecting the most fitting results.
    Our method received 77.78$\%$'s preference while DB+LoRA, TI, ProSpect, CD, ZipLoRA received 2.96$\%$, 4.26$\%$, 0.37$\%$, 1.48$\%$, 13.15$\%$, respectively.
    Results are presented in the right of Fig.~\ref{fig:user_study_ab_test_1} (a), indicating that, overall, our method aligns with both the given content and style reference images simultaneously.
    \input{Figures/optimal_d}
    \item \emph{User Study IV.}
    We conduct A/B testing to evaluate the success rate of generating content-style customized images between our method and other baseline methods.
    During the test, one of the five baseline methods is randomly selected for comparison with our method. Both methods generate nine images with different seeds. Participants were asked to select which set of nine images contained more content-style customized generated images.
    Our method received 96.67$\%$'s preference while compared with DB+LoRA, 94.44$\%$'s preference while compared with TI, 92.22$\%$'s preference while compared with ProSpect, 95.56$\%$'s preference while compared with CD, and 85.56$\%$'s preference while compared with ZipLoRA, respectively.
    Results are shown in Fig.~\ref{fig:user_study_ab_test_1} (b)
    The results indicate that our method generates a greater number of content-style customized images compared to other methods, suggesting a higher success rate in satisfied image generation.
    \item \emph{User Study V.}
    Similar to the experiment settings in User Study IV, participants were tasked with selecting which set of nine images exhibited stronger consistency among them. Our method received 92.22$\%$'s preference while compared with DB+LoRA, 87.78$\%$'s preference while compared with TI, 85.56$\%$'s preference while compared with ProSpect, 87.78$\%$'s preference while compared with CD, and 81.11$\%$'s preference while compared with ZipLoRA, respectively.
    The results indicate that our method generates a more significant number of content-style customized images compared to the other methods, suggesting a higher success rate in satisfied image generation.
    Results are shown in Fig.~\ref{fig:user_study_ab_test_1} (c)
    The results demonstrate that the images generated by our method exhibit stronger consistency among them, indicating that our method, comparatively, has the highest level of stability.
\end{itemize}

\input{Figures/ablate_compare}

\subsection{Ablation Study}
\label{sec:ablation}
\paragraph{The Optimal Dimension $d$ for the Fixed Parameters.}
In Section~\ref{sec:method}, we introduce the hyperparameter $d$ as the row dimension of the fixed parameters, representing the proportion of fixed parameters in the parameter subspace. 
We conduct experiments on eight different fixed parameter ratios, $1/8$, $1/4$, $3/8$, $1/2$, $5/8$, $3/4$, $7/8$, and $1$, corresponding to the proportion of fixed parameters relative to the total parameters.
We quantitatively evaluate the text alignment, content alignment, and style alignment metrics for various values of $d$, and the results are presented in Fig.~\ref{fig:optimal_d}.
From the histogram, we observe that as the ratio of fixed parameters increases, both the values of ``Style Alignment'' and ``Text Alignment'' gradually rise, reaching their peaks at a ratio of $0.5$, and then gradually decline. 
The ``Average Alignment'' reaches its maximum at a ratio of $0.5$.
This indicates that the optimal alignment occurs at a ratio of $0.5$, resulting in better customized content-style images. This finding aligns with our theoretical framework introduced in Section~\ref{sec:method}, where a 1:1 ratio between fixed and trainable parameters results in the ``content parameter subspace'' in Eq.~(\ref{eq:W_dl}) and ``style parameter subspace'' in Eq.~\ref{eq:W_ur}) having the maximum number of trainable parameters, thus reaching the maximum learning capacity and achieving the best generation effect.
It is noteworthy that at a ratio of $0.5$, the ``Content Alignment'' is not maximal. 
This is because the results of other ratios present a weaker learned style (as indicated by lower ``Style Alignment'' in the histogram) and are entangled with the content to some extent.

\paragraph{Multi-Correspondence Projection Learning.}
To prevent overfitting between specific content and style during the training stage, we introduce a \textbf{Multi-Correspondence Projection} (``\textbf{MCP}'') learning within our work.
We conduct two ablation studies with different experimental settings to evaluate the impact of the proposed MCP.
Specifically, in the first study, we train the specific content (e.g., ``vase") with a particular style (e.g., ``glass style'') in a one-to-one manner and leverage the trained model for inference. The results are shown as \textbf{W/o MCP-I} in Fig.~\ref{fig:ablate_compare}.
In the second study, we first train the model on a specific content (e.g., ``vase") with a different style (e.g., ``yarn style'') in a one-to-one manner. Subsequently, in the second stage, we combine the content adapters with the trained style (e.g., ``glass style'') adapters and utilized the combined model for inference. The results are shown as \textbf{W/o MCP-II} in Fig.~\ref{fig:ablate_compare}.
We can observe that in the first study, the generated images exhibit a degree of overfitting to the reference images(e.g., ``teapot'' with ``chair legs'' from the style reference image, ``sticker'' style with realistic style from the content reference image), resulting in a decrease in fidelity to the content or style, thereby reducing the quality of the outputs.
In the second study, we can observe that the results exhibit some features (e.g., the color from ``yarn style'') of the style trained in the first stage. As the final model does not incorporate this style, this is mainly because due to the fact that without MCP, the ``yarn style'' influences the parameter space of the content during the training stage, as analyzed in Fig.~\ref{fig:Multi-correspond}.
We present the results of our methods in \textbf{With MCP}; by comparing, we can observe that with MCP, we can effectively avoid overfitting and generate images with more disentangled content and style.

\input{Figures/ablate_compare_orth}

\paragraph{Orthogonal Fixed Parameters.}
To demonstrate the effectiveness of the orthogonal fixed parameters designed to enhance the content and style fidelity of generated images, we conduct an experiment where we remove the orthogonal fixed parameters and replace them with randomly fixed parameters.
We present the results in Fig.\ref{fig:ablate_compare_orth} for comparison. Without the orthogonality of the fixed parameters, it leads to decreased fidelity for the generated images. 
For instance, in the case of ``vase'', ``teapot'', and ``teddy bear'', the generated images no longer preserve the original content details, and the style has also changed. In the case of the ``sticker style'', the generated images lose the cartoonish style of the contents present in the reference.
We also present quantitative results in Table.~\ref{tab:alignment}. After ablating fixed parameter orthogonalization, although the content alignment slightly increases, the style alignment decreases significantly, and the average alignment decreases as well.
Note that the slight increase in text alignment is due to the increase in content alignment, as the prompts' emphasis on describing the image content.

\paragraph{Fine-Tuning of the Combined LoRA Modules.}
In the second stage of our pipeline, we reconstruct the entity parameter space by combining the content and style PLP matrices.
Subsequently, we fine-tune the combined LoRA modules for a few dozen steps to enable the model to generate images with customized content with style.
The results of ablating the fine-tuning step are presented in Fig.~\ref{fig:fig_reconstruct}. The results demonstrate that after undergoing a few dozen fine-tuning steps, our proposed method achieves optimal visual performance.
We also present individual content or style generation results in the middle and bottom rows of Fig.~\ref{fig:fig_reconstruct}. These results illustrate that our proposed method successfully disentangles content and style while retaining the capability to faithfully generate individual content or style.

\paragraph{Concepts learning ablation.}
We aim to evaluate the learning effect of the desired content or style in comparison with the baseline stable diffusion model.
To achieve this, we employ pseudo words for training and inference of specific content and style.
For the purpose of comparison, we describe content and style using prompts for generation.
As the results presented in Fig.~\ref{fig:ablation_learning}, solely relying on prompts to describe the desired content or style, without learning these representations, fails to capture detailed features from reference images, leads to unfaithful generation of the customized content and style.

\input{Figures/ablation_learning}

\section{Applications and Discussions}
\input{Figures/application_texture.tex}
\input{Figures/application_portrait.tex}

We demonstrate the effectiveness and versatility of our technique across various applications, including content-style customization of diverse textures and portraits.

\paragraph{Application-I: Content-Style Customization of Various Textures.}
Our technique enables the synthesis of high-quality content with a wide variety of user-controlled textures and materials, which can be leveraged for customized product visualization, digital content creation, or material design applications.
We present results for different textures (knit texture, burlap texture, denim texture, and fabric texture) in Fig.~\ref{fig:application_texture}.
The visualized results indicate that our method is capable of customizing generation for a diverse range of textures while maintaining content consistency with the reference images.
With our approach, designers can easily showcase their products with custom material and textile options tailored to customer preferences.
Compared to traditional rendering pipelines requiring extensive modeling and material setup, our data-driven approach significantly streamlines this process.

\paragraph{Application-II: Content-Style Customization of Portraits.}
Another compelling application of our technique is enabling users to generate stylized portraits adhering to diverse artistic styles and visual domains. This capability opens up new creative avenues for digital artists, as well as opportunities in areas like virtual production and AI-assisted artwork creation.
For digital artists and creative professionals, our framework efficiently synthesizes portrait imagery in various artistic styles with fine user control. Fig.~\ref{fig:application_portrait} illustrates examples where we tasked artists to create stylized portraits using our approach in styles like sticker, watercolor painting, and flat cartoons. 
Compared to manual digital painting, our approach dramatically accelerates this creative process while still allowing users to guide stylistic aspects and maintain consistent facial identities.
A key advantage of our approach is its ability to generalize stylized portrait synthesis across numerous visual domains while still allowing users to control diverse scenes, poses, etc.

\paragraph{Bad Cases.}
While our proposed method demonstrates considerable promise in addressing customized content and style fusion, it is essential to acknowledge instances where the model occasionally exhibits an influence from the style reference on the background regions of the generated images, as presented in Fig.~\ref{fig:badcase}.
This observed influence on the background regions in the generated results can be attributed to the limited diversity among the style reference images. 
To mitigate this issue, future iterations of our method will incorporate rigorous regularization techniques and enhanced data preprocessing methodologies.
Furthermore, the integration of cross-validation procedures and model simplification strategies will be explored to promote improved generalization performance.

\input{Figures/badcase}

\paragraph{Limitations.}
While our method performs well on content-style customization, generating images with complex or rare content/style solely by using textual prompts remains a challenging task.
Specifically, our method leverages the class priors in the T2I model when learning the content or style of given images ($e.g.,$ ``a [c1] dog'' leverages ``dog'' as a class prior, ``a [s1] yarn style'' leverages ``yarn'' as a class prior)~\cite{ruiz2023dreambooth}. 
When the customized content or style images are highly complex or rare, obtaining accurate priors through simple prompts becomes challenging, leading to a decrease in the fidelity of the generated images.

%% file: Figures/fig_main_compare.tex
\begin{figure*}[!h]
  \centering
  \includegraphics[width=0.97\linewidth]{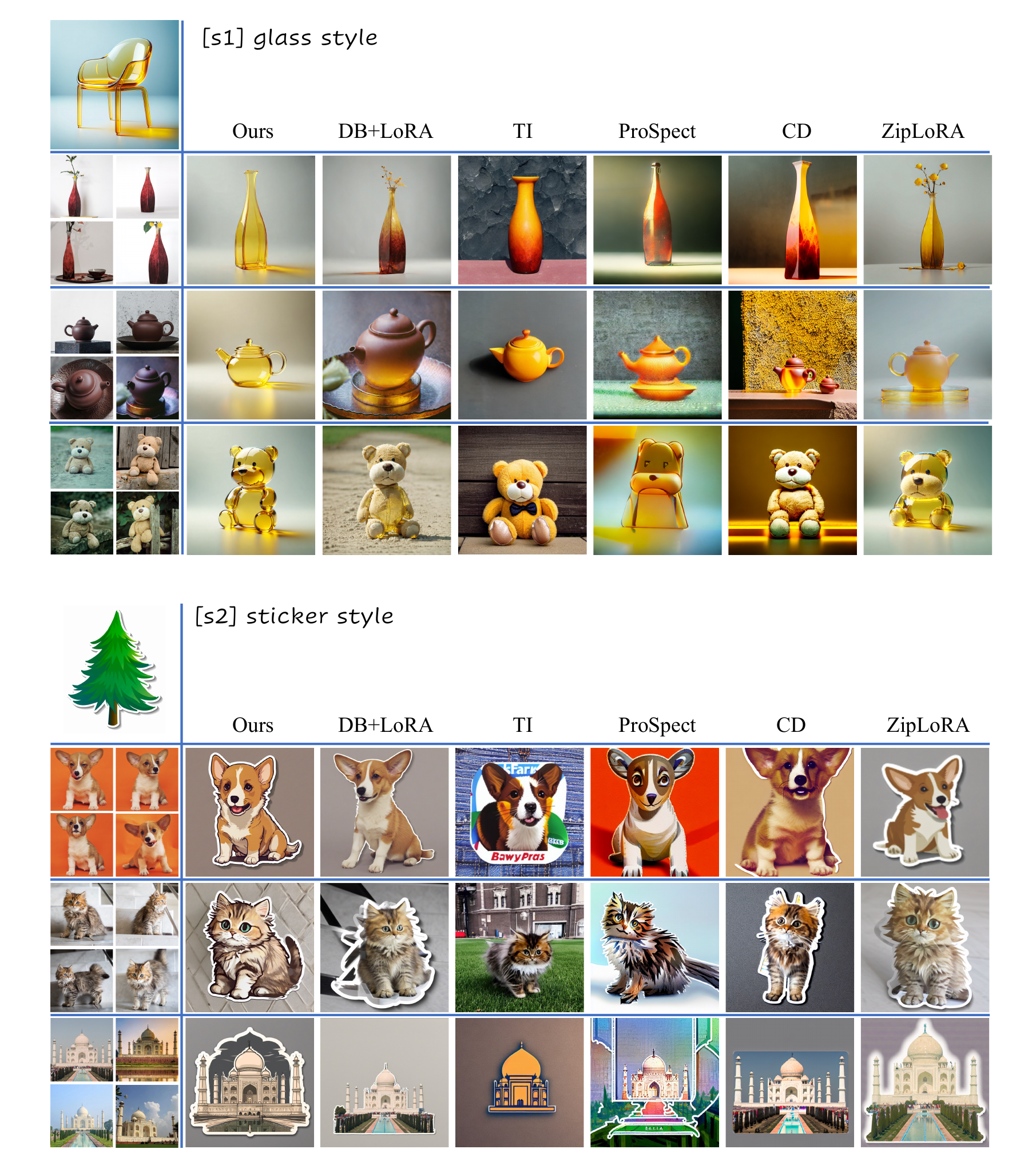}
  \caption{\textbf{Qualitative evaluation and comparison of DB+LoRA, TI, ProSpect, CD, ZipLoRA, and our method in diverse contents.} The results indicate that our method generates harmonious content-style fusion images with diverse contents while preserving the disentanglement of content and style, as well as maintaining high-level fidelity.}
  \label{fig:fig_main_compare}
\end{figure*}

%% file: Figures/first_fig.tex
\begin{figure*}[!th]
\centering
   \includegraphics[width=1\linewidth]{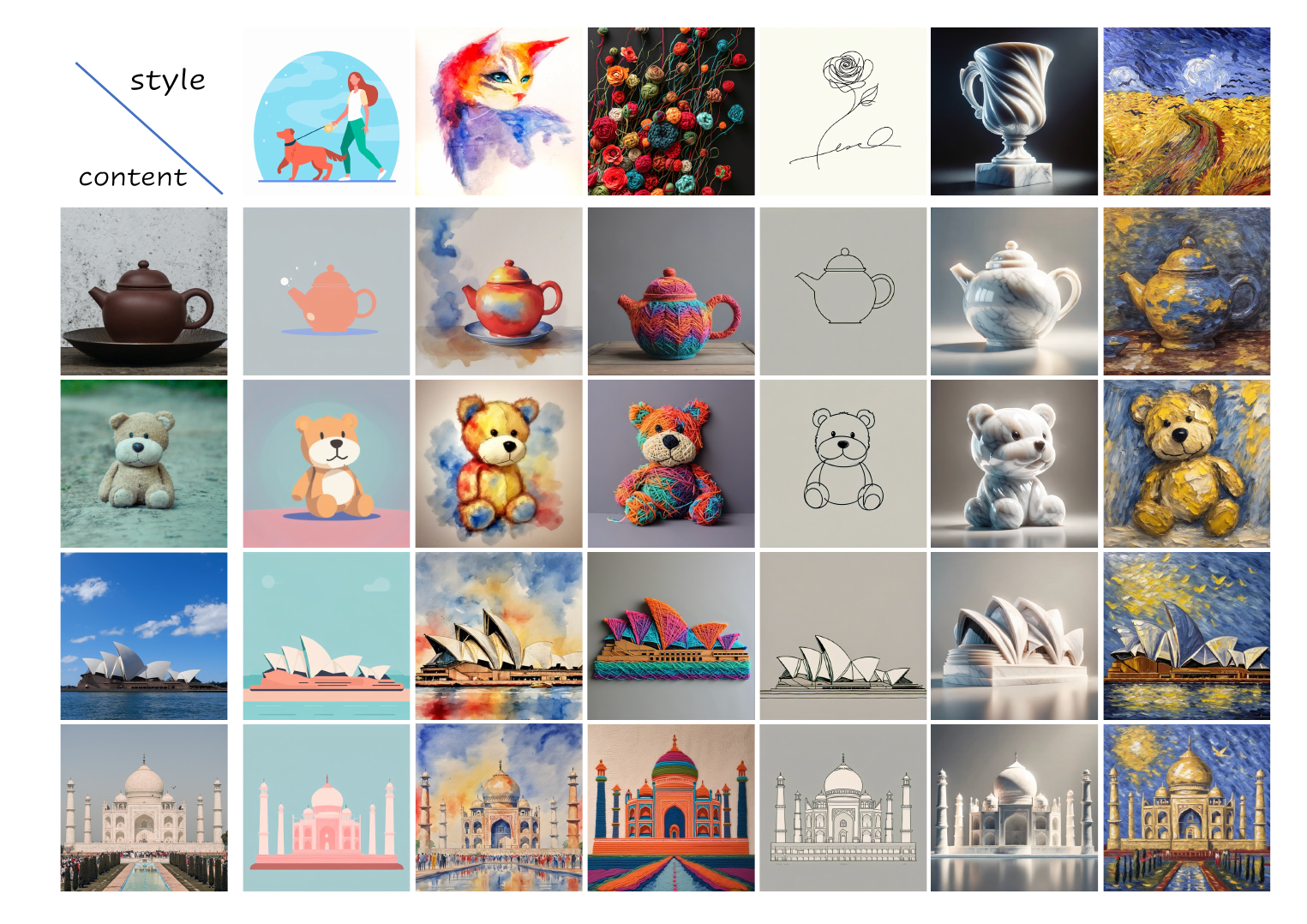}
  \caption{\textbf{More results of diverse content and style generated by our method.}}
  \label{fig:first_fig}
\end{figure*}

%% file: Tables/alignment.tex
\begin{table*}[!th]
\centering
\caption{\textbf{Comparison of cosine similarity between CLIP(for style and prompt) and DINO features(for content) of the generated images and reference style, content, and prompt, respectively}. Our method has the best average score, indicating that our approach successfully customizes the generation of the target content and style while aligning with the prompt.}
\label{tab:alignment}
\begin{tabular}{c|cccccccc}
\hline
Methods                          &DB+LoRA &TI     &ProSpect &CD     &ZipLoRA &w/o MCP &w/o Orth &Ours     \\ \hline
Content-alignment $(\uparrow)$   &0.7982 &0.7292  &0.6165   &0.6845 &0.7103 &0.6242  &\textbf{0.6874}    &0.6615\\ \hline
Style-alignment $(\uparrow)$     &0.4974 & 0.3942 & 0.4816  &0.4381 &0.5414 &0.5331  &0.5626   &\textbf{0.6219} \\  \hline
Prompt-alignment $(\uparrow)$    &0.3894 & 0.2836 & 0.3156  &0.2778 &0.3319 &0.3867  &\textbf{0.4035}    &0.3908\\ \hline
Average $(\uparrow)$             &0.5617 & 0.4690 & 0.4712  &0.4668 &0.5279 &0.5147  &0.5512    &\textbf{0.5581}\\ \hline
\end{tabular}
\end{table*}

%% file: Figures/fig_editability.tex
\begin{figure*}[ht!]
  \centering
  \includegraphics[width=1\linewidth]{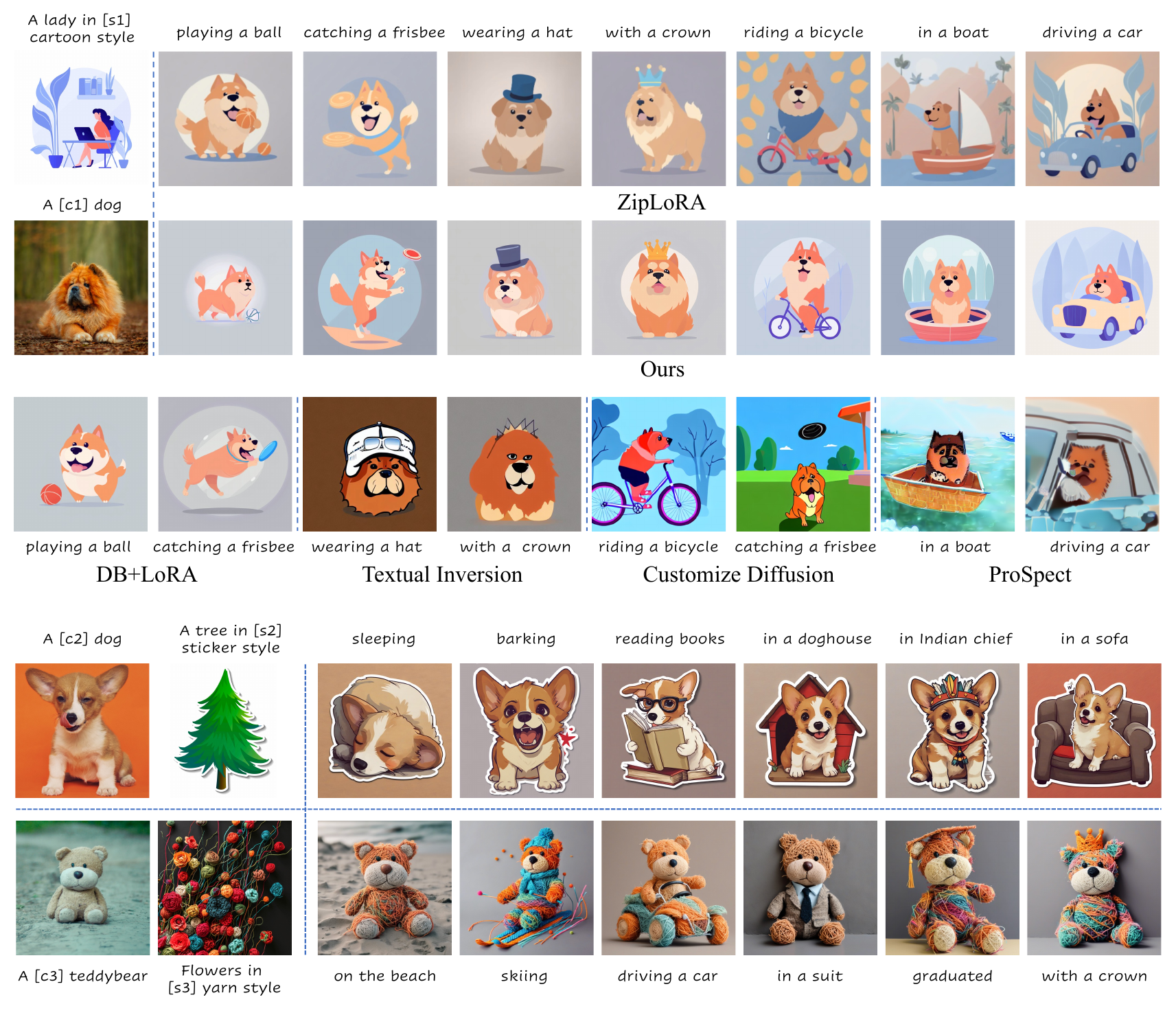}
  \caption{\textbf{Results of generating diverse customized content-style images.} This indicates that our method exhibits excellent editing capabilities as well as generalization capabilities to both content and style.}
  \label{fig:fig_editability}
\end{figure*}

%% file: Figures/tsne.tex
\begin{figure}[!th]
  \centering
  \includegraphics[width=1\linewidth]{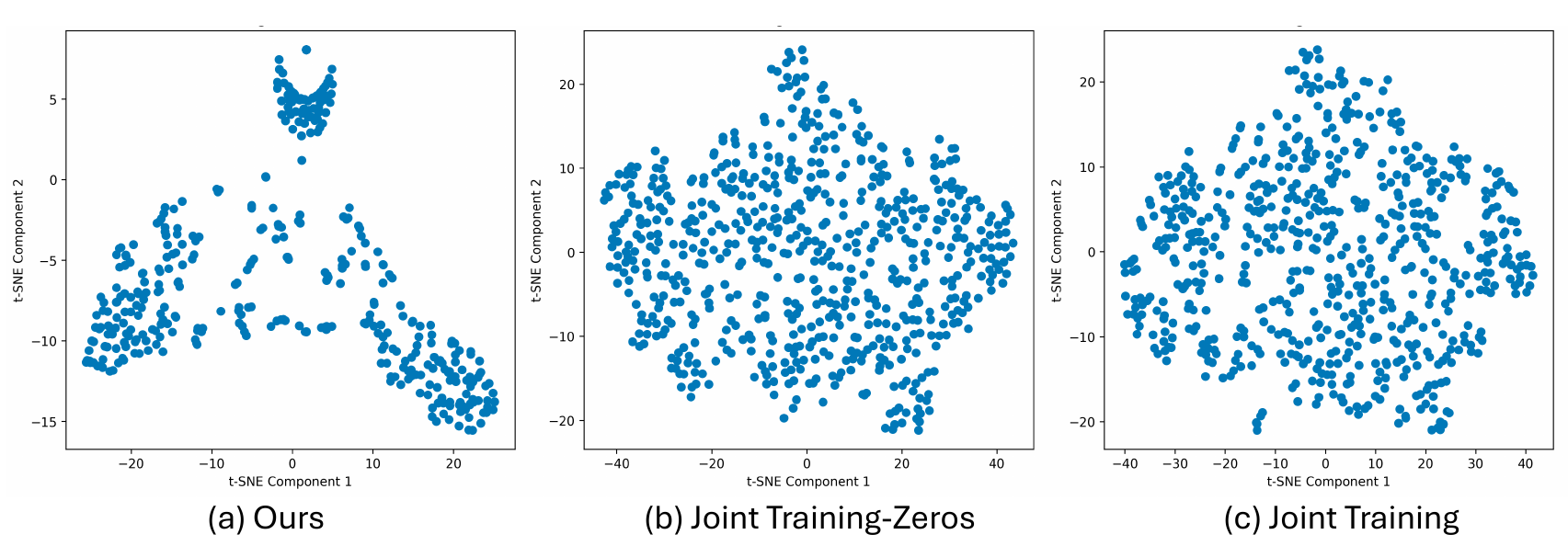}
  \caption{\textbf{Visualizing Low-Rank Adapter Parameter Distributions via t-SNE.}}
  \label{fig:tsne}
\end{figure}

%% file: Figures/fig_reconstruct.tex
\begin{figure}[t!]
  \centering
  \includegraphics[width=1\linewidth]{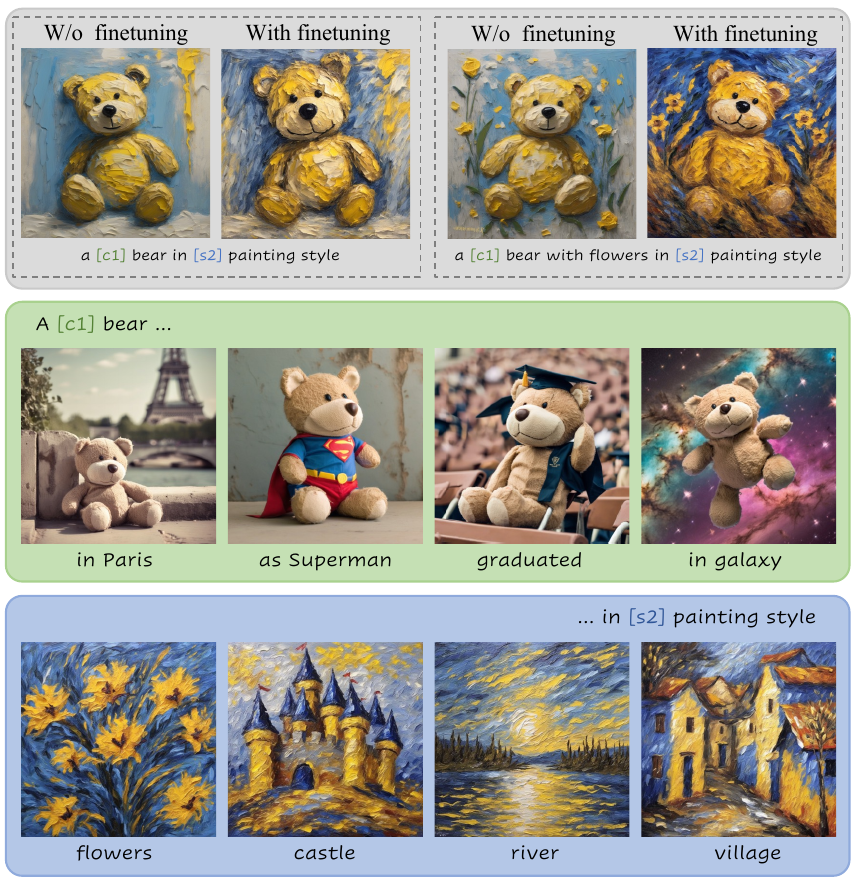}
  \caption{\textbf{Individual content/style generation of our method.} Our method can generate diverse content/style images individually with a high level of fidelity and disentanglement. Fine-tuning enhances the final effect.}
  \label{fig:fig_reconstruct}
\end{figure}

%% file: Figures/compare_edit.tex
\begin{figure}[!t]
  \centering
  \includegraphics[width=1\linewidth]{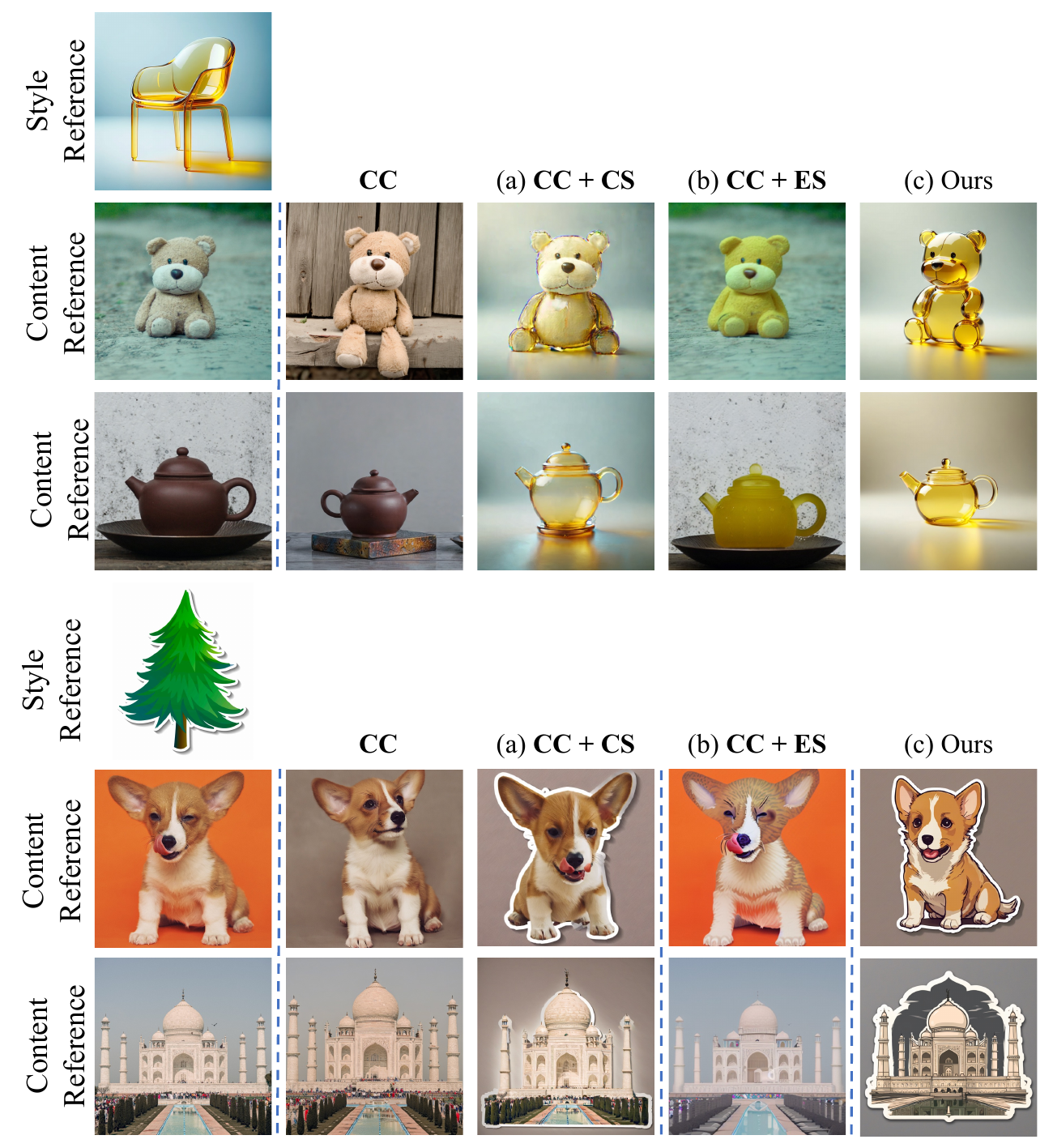}
  \caption{\textbf{Comparison with other two-stage content-style customization paradigms.} \textbf{CC} indicates custom content in the first stage, \textbf{CC+CS} indicates  custom style in the second stage based on \textbf{CC}. \textbf{CC+ES} indicates editing style based on \textbf{CC}.}
  \label{fig:compare_edit}
\end{figure}

%% file: Figures/user_study_abtest_1.tex
\begin{figure}[ht!]
  \centering
  \includegraphics[width=1\linewidth]{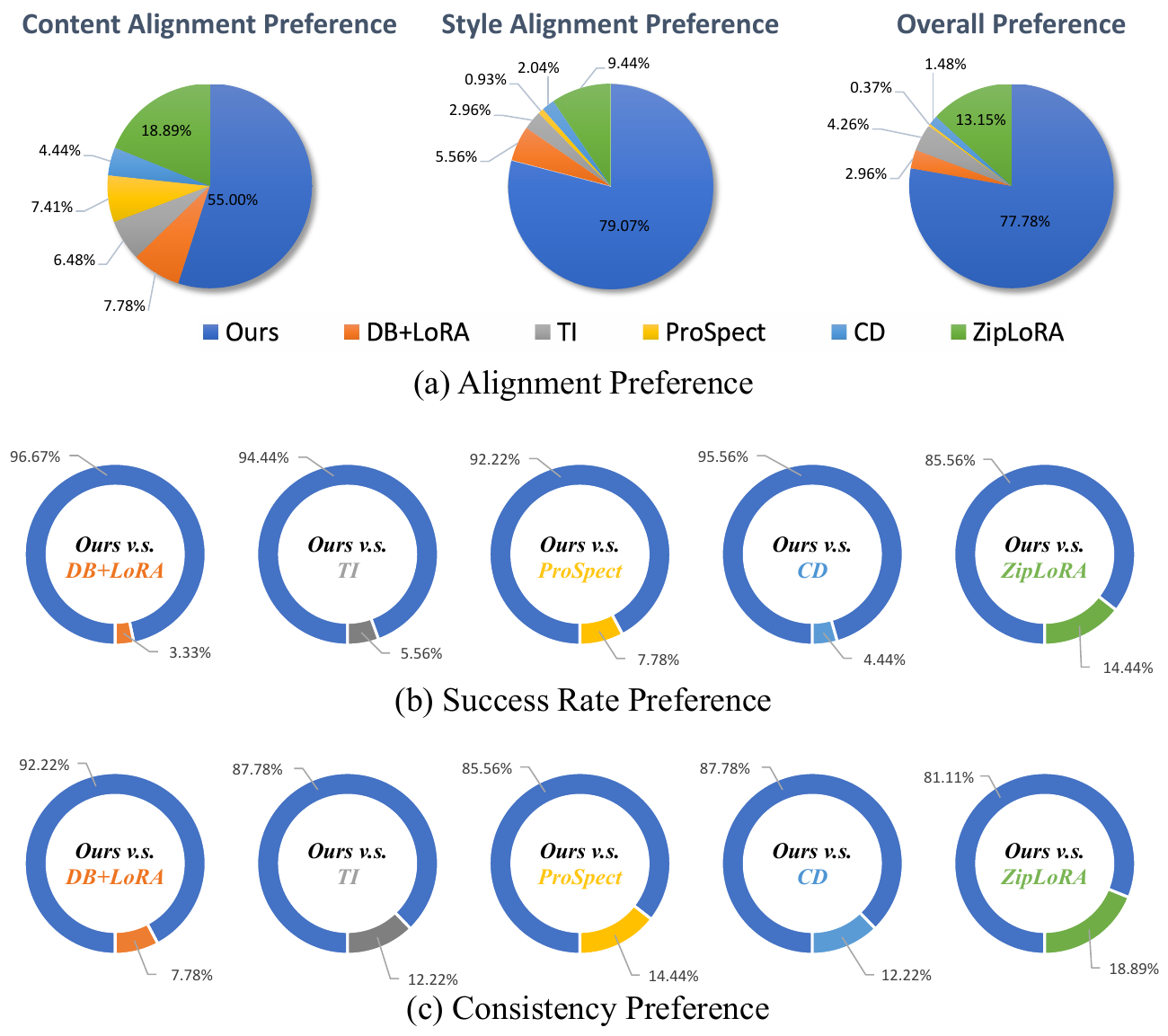}
  \caption{\textbf{User study results.}}
  \label{fig:user_study_ab_test_1}
\end{figure}

%% file: Figures/optimal_d.tex
\begin{figure}[t!]
  \centering
  \includegraphics[width=1\linewidth]{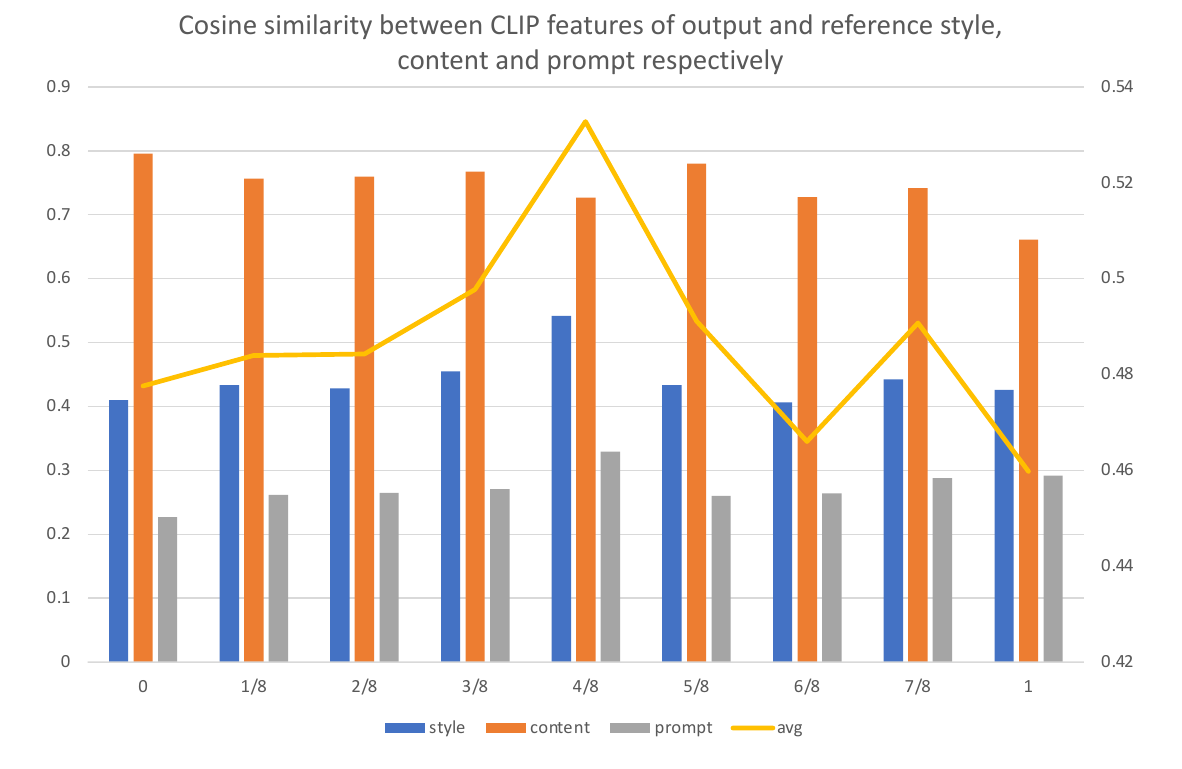}
  \caption{\textbf{Cosine similarity between features of output and reference style,
content and prompt of different dimension d.} When d=0.5, the average cosine similarity of the features reaches its maximum, indicating optimal alignment between the generated results and the reference content, style, and prompt.}
  \label{fig:optimal_d}
\end{figure}

%% file: Figures/ablate_compare.tex
\begin{figure*}[ht!]
  \centering
  \includegraphics[width=1\linewidth]{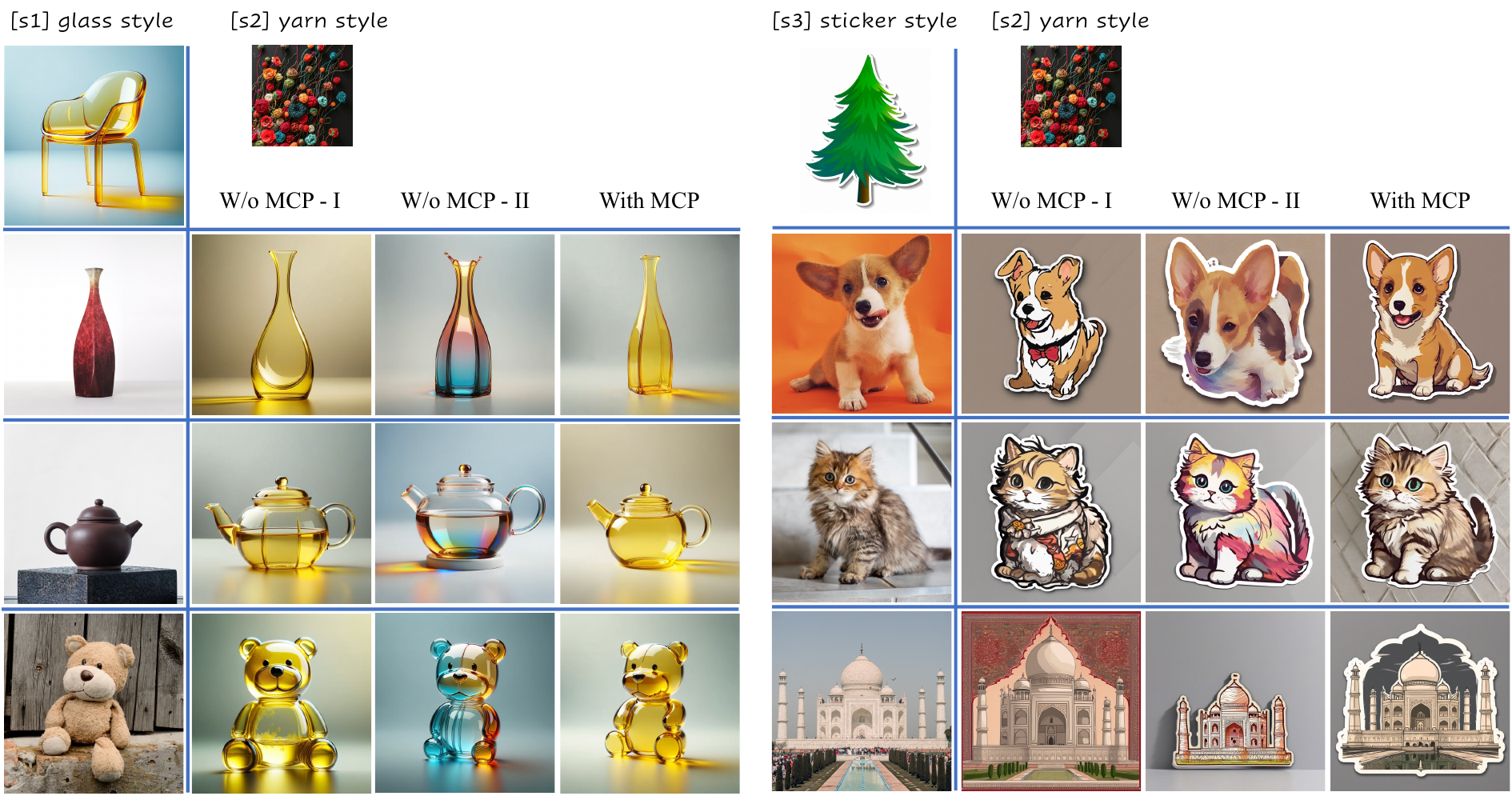}
  \caption{\textbf{Ablation study evaluating the impact of the proposed Multi-Correspondence Projection (``MCP'')}. 
  We train specific content (e.g., ``vase'') and style (e.g., ``glass'') in a one-to-one manner and directly inference after training. Results are presented in \textbf{W/o MCP - I} column.
  We train specific content (e.g., ``vase'') and style (e.g., ``yarn'') in a one-to-one manner in the first stage, and combine the content (e.g., ``vase'') adapters with other style (e.g., ``glass'') adapters in the second stage, then inference with the combined adapters. Results are presented in \textbf{W/o MCP - II} column.
  The visual comparison highlights the effectiveness of MCP in enhancing the details while preserving the disentanglement of content and style, as well as maintaining high-level fidelity of them.}
  \label{fig:ablate_compare}
\end{figure*}

%% file: Figures/ablate_compare_orth.tex
\begin{figure}[t!]
  \centering
  \includegraphics[width=1\linewidth]{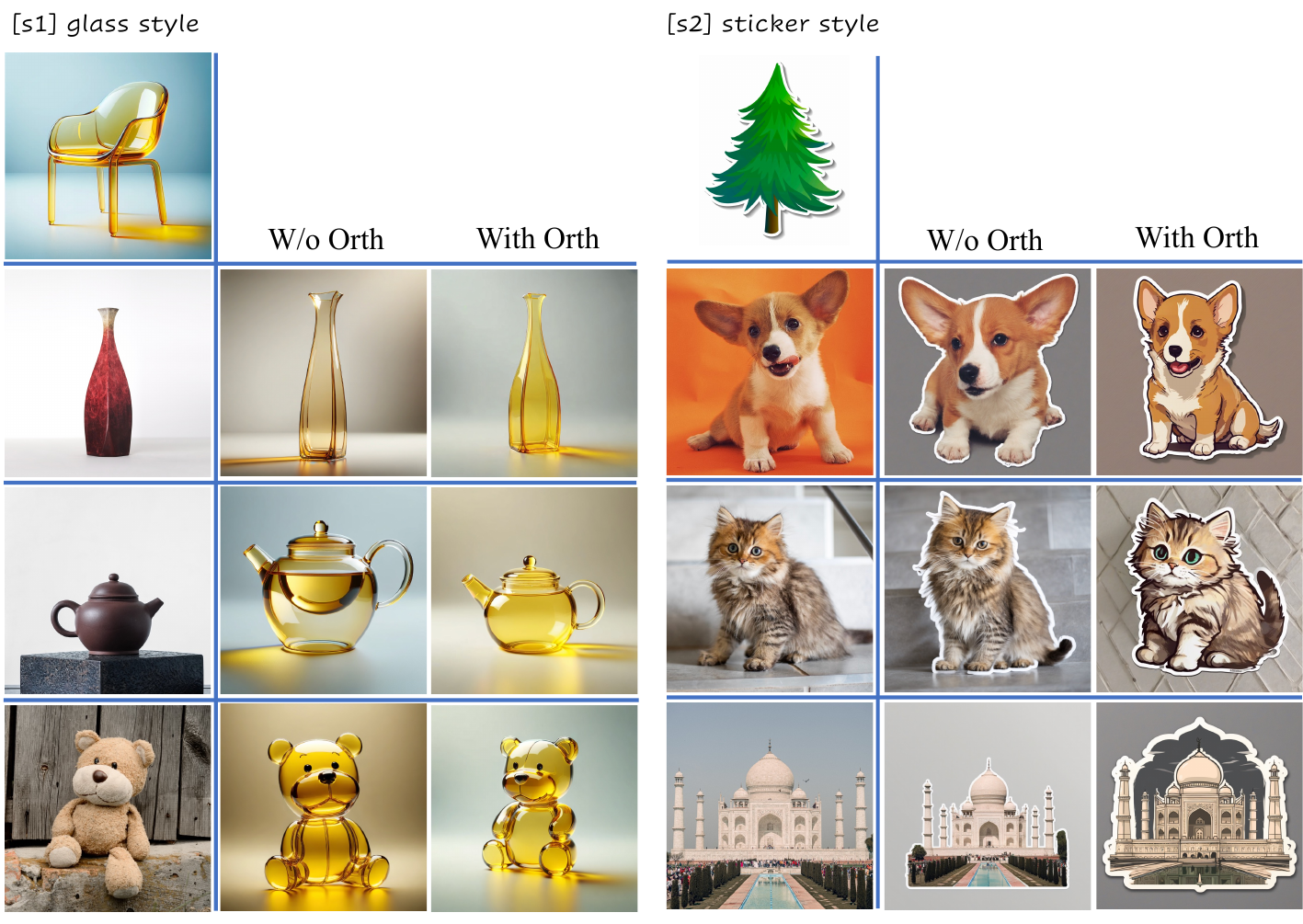}
  \caption{\textbf{Ablation study evaluating the impact of the proposed orthogonal fixed parameters}.
  The \textbf{W/o Orth} shows results without orthogonal fixed parameters, while the \textbf{With Orth} demonstrates the improved image quality achieved by our full method incorporating orthogonal fixed parameters.
  The visual comparison highlights the effectiveness of orthogonal fixed parameters in enhancing content and style fidelity of generated images.}
  \label{fig:ablate_compare_orth}
\end{figure}

%% file: Figures/ablation_learning.tex
\begin{figure}[t!]
  \centering
  \includegraphics[width=1\linewidth]{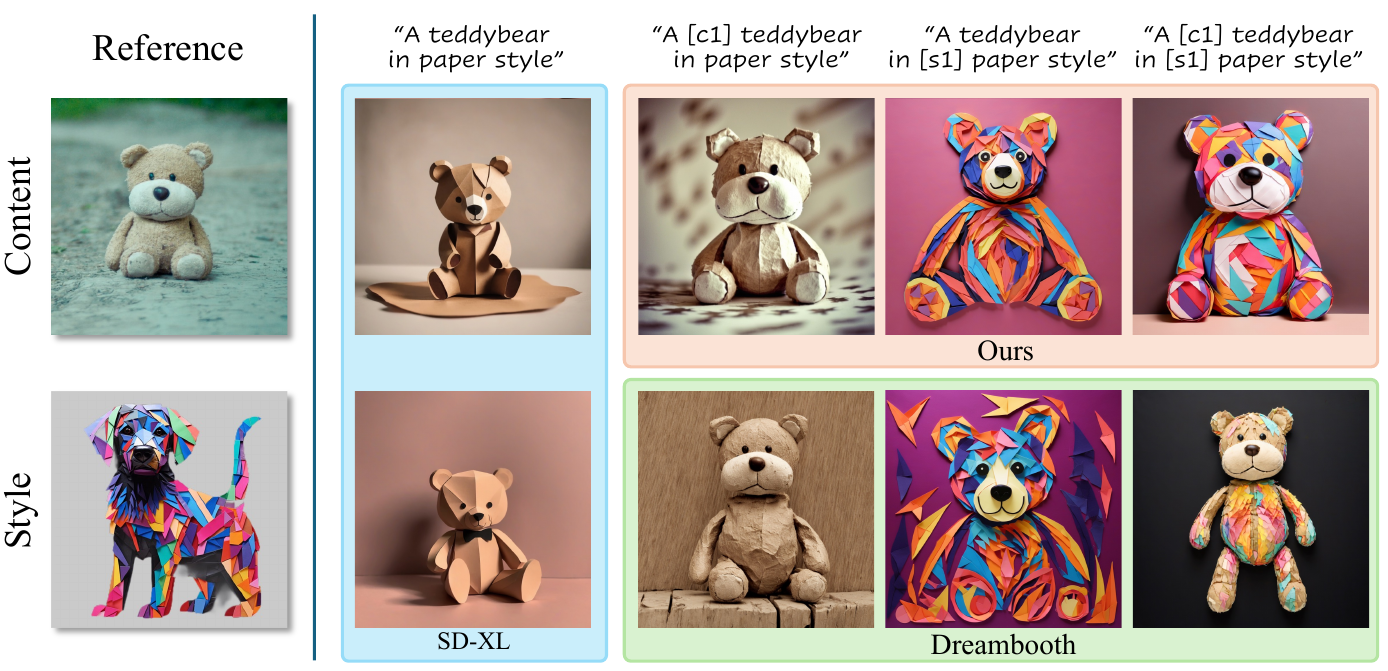}
  \caption{\textbf{Comparison of results with and without learning concepts.}
   We present output images generated with and without learning reference content or style in in the orange (by our method) and green (by DreamBooth method) boxes.
   We also show images directly generated by basic Stable Diffusion-XL model in blue box.
   Prompts for inference are shown on top. 
   Without learning content or style in pseudo words, models that rely solely on prompts cannot generate desired content or styles faithfully.
   }
  \label{fig:ablation_learning}
\end{figure}

%% file: Figures/application_texture.tex
\begin{figure}[t!]
  \centering
  \includegraphics[width=1\linewidth]{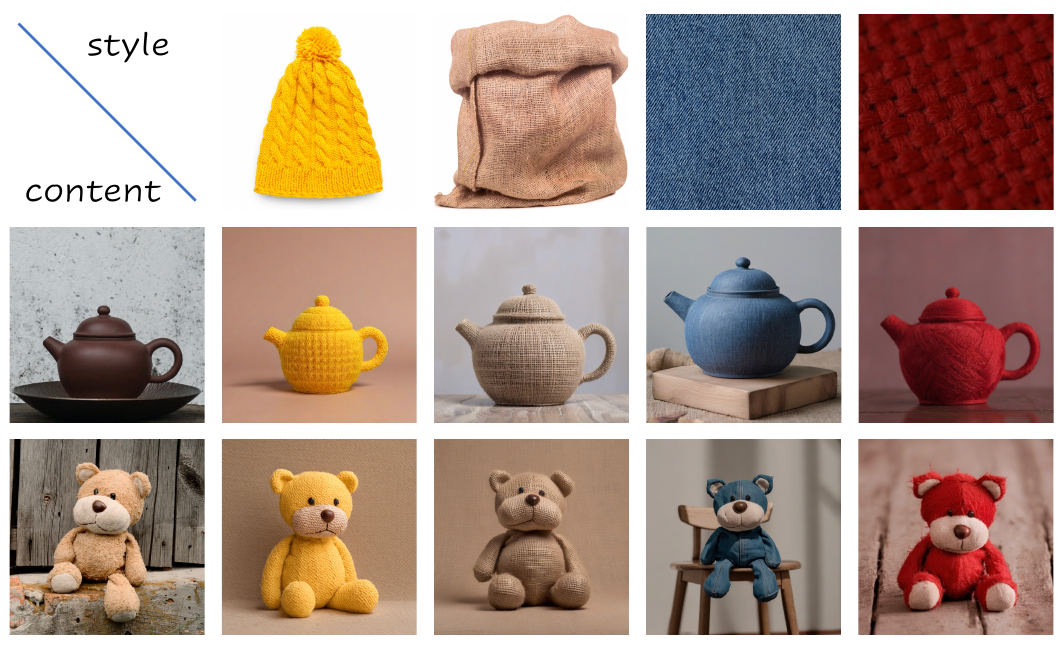}
  \caption{\textbf{Application I.} Content-style customization of variety texture, including knit, burlap, denim and fabric texture.}
  \label{fig:application_texture}
\end{figure}

%% file: Figures/application_portrait.tex
\begin{figure}[t!]
  \centering
  \includegraphics[width=1\linewidth]{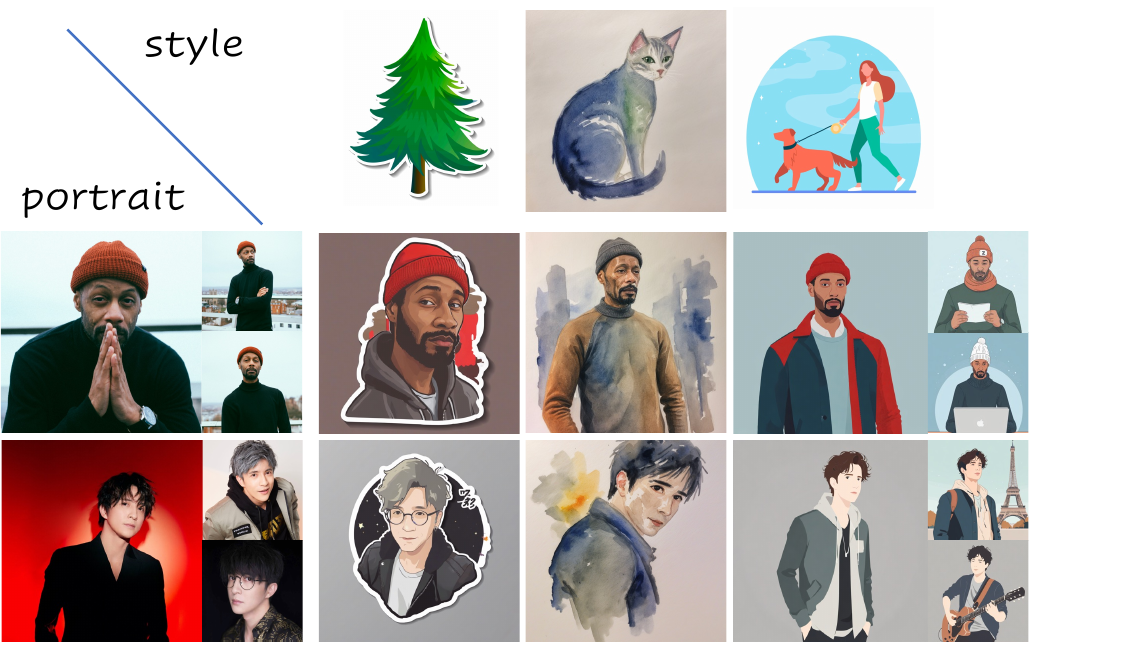}
  \caption{\textbf{Application II.} Content-style customization of portraits. Image credits:@Philip Martin~\cite{Philip2023unsplash}(up)}
  \label{fig:application_portrait}
\end{figure}

%% file: Figures/badcase.tex
\begin{figure}[t!]
  \centering
  \includegraphics[width=1\linewidth]{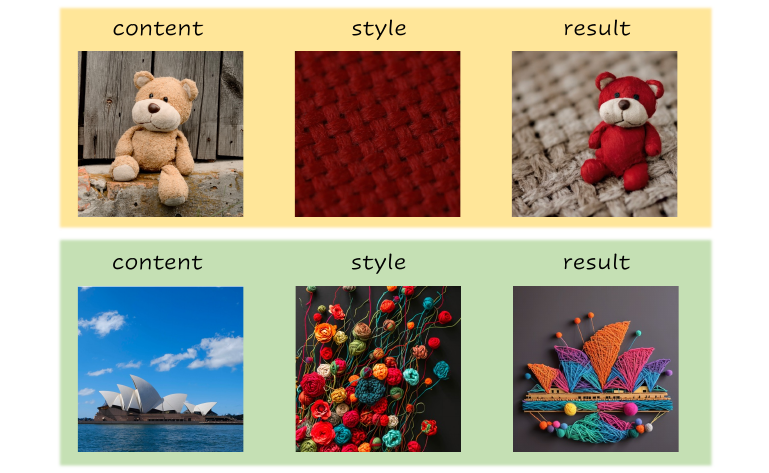}
  \caption{\textbf{Bad cases of our method.} Sometimes style reference can induce undesirable influences on the background generation in the output results.}
  \label{fig:badcase}
\end{figure}